\documentclass{bmvc2k}

\usepackage{amsmath}
\DeclareMathOperator*{\argmax}{arg\,max}

\usepackage{dsfont}
\usepackage{sidecap}
\usepackage{floatrow}

\newenvironment{smitemize}{
\begin{itemize}
  \setlength{\topsep}{-3pt}
  \setlength{\itemsep}{1pt}
  \setlength{\parskip}{0pt}
  \setlength{\parsep}{0pt}
  \setlength{\leftmargin}{1em}  
  \setlength{\itemindent}{-4pt}
}{\end{itemize}}

\newcommand{\paragraphX}[1]{\vskip 0.2cm \noindent \textbf{#1} \hskip .1cm}
\newcommand{\squeezeup}{\vspace{-3mm}}

\usepackage{multirow}
\usepackage{array}
\usepackage{stackengine}
\usepackage{amsfonts}
\usepackage{xcolor}
\usepackage{wrapfig}

\title{Gradient Frequency Modulation for Visually Explaining Video Understanding Models}

\addauthor{Xinmiao Lin}{xl3439@rit.edu}{1}
\addauthor{Wentao Bao}{wb6219@rit.edu}{1}
\addauthor{Matthew Wright}{Matthew.Wright@rit.edu}{1}
\addauthor{Yu Kong}{Yu.Kong@rit.edu}{1}

\addinstitution{
 Golisano College of Computing and Information Sciences \\
 Rochester Institute of Technology\\
 Rochester, NY
}

\runninghead{XINMIAO, WENTAO, MATTHEW, YU}{GFM FOR EXPLAINING VIDEO MODELS}

\begin{document}

\maketitle

\begin{abstract}
In many applications, it is essential to understand why a machine learning model makes the decisions it does, but this is inhibited by the black-box nature of state-of-the-art neural networks. Because of this, increasing attention has been paid to explainability in deep learning, including in the area of video understanding. Due to the temporal dimension of video data, the main challenge of explaining a video action recognition model is to produce spatiotemporally consistent visual explanations, which has been ignored in the existing literature. In this paper, we propose Frequency-based Extremal Perturbation (F-EP) to explain a video understanding model's decisions. Because the explanations given by perturbation methods are noisy and non-smooth both spatially and temporally, we propose to modulate the frequencies of gradient maps from the neural network model with a Discrete Cosine Transform (DCT). We show in a range of experiments that F-EP provides more spatiotemporally consistent explanations that more faithfully represent the model's decisions compared to the existing state-of-the-art methods.
\end{abstract}

\section{Introduction}
Since the first major success of deep learning in 2012, there has been an explosion of applications of these models, such as in image classification~\cite{imagenet}, machine translation~\cite{machine_trans} and tumor detection~\cite{tumor_detection}. As people interact more frequently with these models, and they are applied in critical applications like medical diagnosis~\cite{richens2020improving} and terrorism detection~\cite{detection}, it is becoming increasingly important to understand their decisions. Otherwise, it can be hard for people to trust the models, since blind trust could result in catastrophic consequences. A plethora of evidence shows the importance of explanation towards understanding and building trust in cognitive psychology~\cite{cognitive-psychology}, philosophy~\cite{philosophy} and machine learning~\cite{ml-1, ml-2}. 

Numerous works have emerged, especially in the computer vision field, to explain the decisions of deep neural networks. Researchers have come to a consensus that model explanation should be interpretable and faithful to the model~\cite{xai, grad-cam, integrated}. A common type of model explanation in the image domain takes the form of a heatmap or {\emph saliency map} that localizes the salient areas of an input for a model's decision. Among these approaches, perturbation-based methods~\cite{meaningful, extremalPerturbations, FGVis} produce explanations that are more faithful and fine-grained than CAM- and backpropagation-based methods~\cite{cam, grad-cam, guided-backprop, lrp}, but also produce noise that appears to humans as randomly selected, which hurts their interpretability. See Fig.~\ref{fig:comparison_vis} for a comparison of these explanation methods.

Another challenge arises when the aforementioned methods for the image domain are leveraged to the video domain: the explanations fail to capture the motion dynamics. We argue that \emph{spatiotemporal consistency} should be added as another criterion for model explanation in the video domain. A spatiotemporally consistent explanation should focus on the target object/scene and follow its spatiotemporal trajectory, such as the gymnast in Fig.~\ref{fig:comparison_vis}. This greatly helps the interpretability of the explanation. At the same time, the video understanding models may indeed not pay attention to the target object/scene at each frame because of the moving edges, shot changes, and other issues~\cite{c3d, r3d}. Spatiotemporally consistent explanations help the end-users or the developers of models to better assess the model's understanding of the dataset and improve upon it. If the explanations are faithful to the model/data, they may not be spatiotemporally consistent; and if they are more spatiotemporally consistent, they may lose faithfulness. Thus, the challenge resides in finding a balance between the faithfulness of an explanation and its interpretability.

Therefore, we propose the Frequency-based Extremal Perturbation method (F-EP) which extends the EP method~\cite{extremalPerturbations} to the video domain. F-EP, illustrated in Fig.~\ref{fig:f-ep}, aims to find the salient areas of the input by perturbing a mask, where the perturbations are the gradients of the output label with respect to the input. The masks, comprised of the aggregated perturbations, become the explanations for the video input. At each iteration of the optimization process, F-EP transforms the gradients into the frequency space using a common Fourier Transform: Discrete Cosine Transform (DCT). Then, F-EP modulates the frequency signals and transforms them back to the gradient space to update the masks.

This approach is inspired from works~\cite{high-frequency, jpeg} showing that neural networks not only learn the low-frequency components of an image, such as faces and shapes, but also the high-frequency components that appear like noise to humans. Therefore, the gradients of the salient high-frequency components will be high and make the gradient maps noisy. To overcome this problem, F-EP proposes to transform the gradients into the frequency domain and selects a combination of low and high frequency components of the gradients to achieve semantically meaningful explanations that are also faithful to the model. Meanwhile, because the explanations focus on the semantic features of each frame, such as a person performing an action, and the majority of frames contain these features, the explanations will naturally become spatiotemporally consistent.

Since no existing metrics address spatiotemporal consistency of  explanations, we propose the \emph{Spatiotemporal Consistency (STC)} metric to measure how well the explanations align with the ground truth bounding boxes. Because bounding boxes indicate the spatial position of the foreground object/scene, spatiotemporal consistent explanations will have high alignment with them. We experimentally show that F-EP achieves superior performance in terms of spatiotemporal consistency and faithfulness than the state-of-the-art methods. In summary, our contributions in this paper are as follows.
\begin{smitemize}
    \item We propose the F-EP method, which can simultaneously reduce the noise in the explanations and make them spatiotemporally consistent by modulating the gradients in the frequency space.   
    \item We propose a new metric, Spatiotemporal Consistency (STC), to accurately assess the spatiotemporal consistency of an explanation in the video domain.
    \item In the ablation studies section, we show that F-EP using low frequency gradients is able to achieve state-of-the-art performance, while the addition of high-frequency gradients can improve its performance even further.
\end{smitemize}

\section{Related Work} 

Model explainability is increasingly popular in a diverse range of deep learning applications. In this section, we present only the methods for image and video recognition models.

\vspace{-5pt}
\paragraphX{Class Activation Map (CAM)-based.} 
CAM-based methods, such as CAM~\cite{cam}, Grad-CAM~\cite{grad-cam}, Grad-CAM++~\cite{grad-cam++}, and Score-CAM~\cite{score-cam}, use weighted sums of either activation maps or gradients of the class label with respect the input to produce a heatmap. Although CAM-based methods are efficient~\cite{ablation-cam, xgrad-cam}, usually requiring just a few forward and backward passes, the resulting heatmaps are coarse due to the upsampling process. The Saliency Tubes approach~\cite{saliency-tubes} adapts Grad-CAM to the video domain.  
    
\vspace{-3pt}
\paragraphX{Backprop-based.} 
Backprop-based methods~\cite{deconvnet,gradient,integrated,smoothGrad,lrp,clrp,excitation,cebr} compute saliency scores for each input pixel by starting with the final layer and, one layer at a time, inferring the importance of the inputs to the outputs of the layer until the input is reached. For example, Layerwise Relevance Backpropagation (LRP)-based methods~\cite{lrp, clrp} assign a relevance score for each neuron and backpropagate the relevance score of each neuron to the input layer. 
The explanations of backprop-based methods have been criticized for being similar to the results of an edge detector~\cite{sanity-checks} and insensitive to randomization of the model's parameters, which should intuitively alter the results~\cite{exp-lie}.
    
\vspace{-3pt}
\paragraphX{Perturbation-based.} 
Perturbation-based explanation methods~\cite{LIME,rise,gradient,meaningful,extremalPerturbations,FGVis,step} add perturbations to the input features to quantify the importance of each pixel. For example, RISE~\cite{rise} measures the importance of pixels by randomly masking areas of input features and observing changes to the output. 
Meaningful Perturbation (MP)~\cite{meaningful} and Extremal Perturbation (EP)~\cite{extremalPerturbations} methods optimize masks as explanations to localize salient areas. 
STEP~\cite{step} extends EP to the video domain by adding spatiotemporal smoothness constraints, and it is the current state-of-the-art for explaining video understanding models. Although perturbation-based methods~\cite{extremalPerturbations, step} are able to produce more fine-grained explanations and are more sensitive to changes in the model's weights compared to the CAM-based and backpropagation-based methods~\cite{grad-cam, grad-cam++, gradient, integrated}, their explanations are still noisy and not spatiotemporally consistent (see Fig.~\ref{fig:comparison_vis}). Our proposed F-EP approach probes the frequency signals in the gradients in order to denoise the explanations and make them focus more tightly on the target object in every frame.

\vspace{-3pt}
\paragraphX{Fourier Transforms.} 
Fourier transforms decompose a data point in the spatial or temporal domain into a combination of functions in the corresponding frequency domain. Discrete Cosine Transform (DCT) is a type of Fourier Transform using only real numbers. DCT was first proposed for efficient image compression, where only the content-defining features are preserved~\cite{jpeg}. DCT is used in machine learning for image compression with SVM~\cite{svm-compress}, image classification~\cite{dct-class}, and more. Recently, \cite{guo-low, low-frequency} explore the effectiveness of using DCT on crafting adversarial samples. Although F-EP is also using perturbation to produce explanations, the perturbations are added \emph{on the masks} and not the inputs as in~\cite{guo-low, low-frequency}. The optimization objectives are also different: adversarial attacks aim to generate samples to change the model's prediction, while explanations aim to faithfully represent a model's decision. Also, the adversarial samples are sparse, noisy and hard to interpret, while the explanations given by F-EP focus tightly on the target object/scene spatially and temporally.

\section{Methodology}
\begin{figure}[t]
    \centerline{\includegraphics[width=1.0\textwidth]{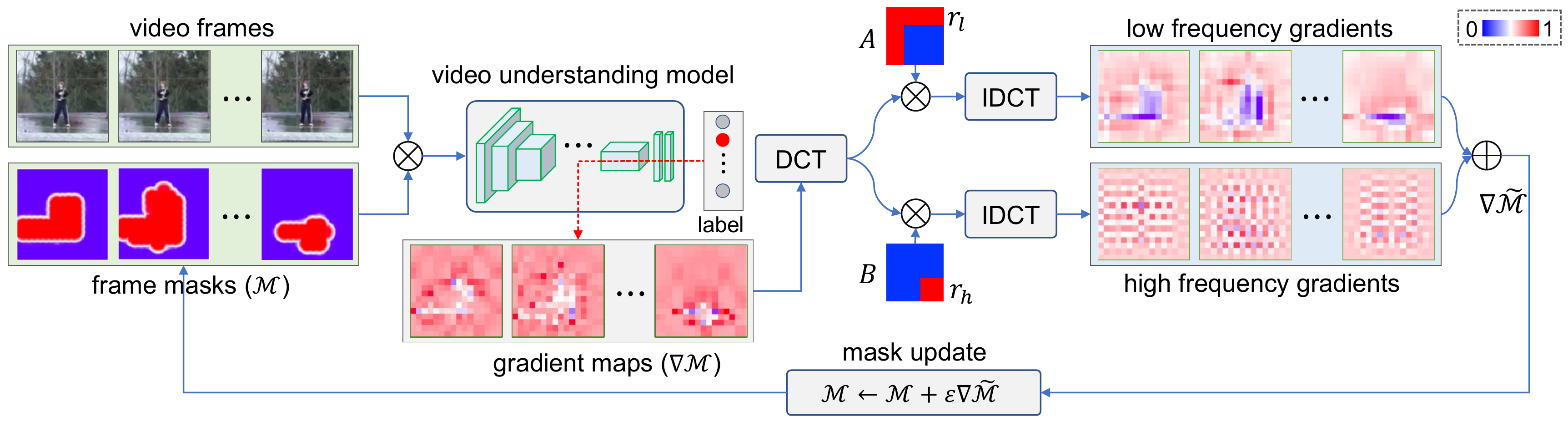}}
    \captionsetup{aboveskip=3pt}
    \caption{\small{Frequency-based Extremal Perturbation (F-EP). The video frames perturbed with masks $M$ are fed to the video understanding model to compute the gradients $\nabla M$. $\nabla M$ is transformed to the frequency domain using DCT. $r_l$ and $r_h$ are the ratios of low- and high- frequency signals to keep in the gradients. The modulated gradients are then transformed back to the video domain using IDCT and combined together to get $(\nabla \Tilde{M})$, which is then used to update the masks.}}
    \label{fig:f-ep}
\end{figure}

In this paper, F-EP is experimented on video classification models as in EP~\cite{extremalPerturbations}. A future research direction could be to evaluate F-EP on other types of video understanding models. Denote the video classification model by $\Phi$, the original input video clip as $X \in \mathbb{R}^{T \times C \times H \times W}$ and the predicted label $y = \Phi(X)$, where $y$ is among a set of $Y$ classes, $T$ is the number of frames in a video clip, $H$ and $W$ are the height and width of the frame/mask, and $C$ is the number of channels. We first briefly introduce the baseline method, Extremal Perturbation (EP)~\cite{extremalPerturbations}, and then explain our proposed Frequency-based Extremal Perturbation (F-EP) method, which is shown in Fig.~\ref{fig:f-ep}.

\subsection{Extremal Perturbation}
Extremal Perturbation (EP) was proposed to explain image understanding models and aims to localize the salient areas in the input for a model's decision through perturbations on the masks. Let the masks be $M \in \mathbb{R}^{T \times 1 \times H \times W}$, the optimization objective of EP is:
\begin{equation}
 \begin{split}
    & M^*_a = \argmax_{M} \Phi_y (M \otimes X) - \lambda R_a(M).
    \label{eq:ep}
 \end{split}
\end{equation}
The left term in Eq.~\eqref{eq:ep} maximizes the classification confidence of the model $\Phi$ to the perturbed input $(M \otimes X)$. The operator $\otimes$ is the perturbation operator which applies local Gaussian blurring to each pixel in $M$. The second term in Eq.~\eqref{eq:ep} is $R_a(M) = \lVert \text{vecsort}(M) - r_a \rVert ^2$, which regularizes the area of the mask $M$ with respect to the area constant $a$. $\lambda$ is the regularizer constant. $r_a$ is a binarized vector with $a$ ones and $(1-a)$ zeros, and $\text{vecsort}(M)$ is sorted vector of the masks $M$. Thus, $R_a(M)$ computes the loss of the area of masks with respect to the area constraint $a$. To solve the optimization problem~\eqref{eq:ep}, we can derive the gradients of the output class label with respect to the masks where $\epsilon$ is the learning rate of the optimization (details in~\cite{extremalPerturbations}):
\begin{equation}
 \begin{split}
    & \nabla M = \frac{\partial(\Phi_y(M \otimes X) - \lambda R_a(M))}{\partial M}, 
    \label{eq:gradients-ep}
 \end{split}
\end{equation}
The optimal solution $M^*$ thus can be approximated by an iterative gradient ascent process. In each iteration, the masks $M$ are updated as follows:
\begin{equation}
    \begin{split}
        M^{f+1} = M^{f} + \epsilon \nabla M,
        \label{eq:gradient-ascent}
    \end{split}
\end{equation}

\subsection{Frequency-based Extremal Perturbation}

\newcommand{\curvefigwidth}{0.092\textwidth}

\begin{figure*}
\setlength{\abovecaptionskip}{0.0cm}
\setlength{\belowcaptionskip}{0.1cm}
    \begin{minipage}[b]{1.0\linewidth}
      \subfigure[EP~\cite{extremalPerturbations} Gradient Maps]{
        \includegraphics[width=\curvefigwidth]{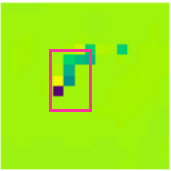}
        \includegraphics[width=\curvefigwidth]{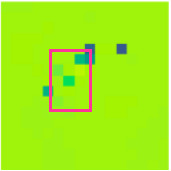}
        \includegraphics[width=\curvefigwidth]{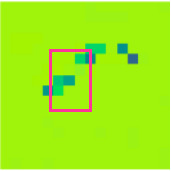}
        \includegraphics[width=\curvefigwidth]{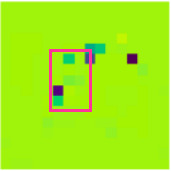}
        \includegraphics[width=\curvefigwidth]{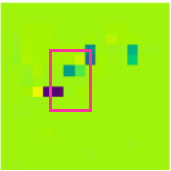}
      }%
      \centering
      \subfigure[F-EP Gradient Maps]{
        \includegraphics[width=\curvefigwidth]{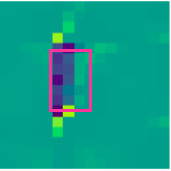}
        \includegraphics[width=\curvefigwidth]{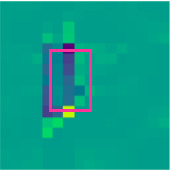}
        \includegraphics[width=\curvefigwidth]{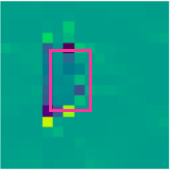}
        \includegraphics[width=\curvefigwidth]{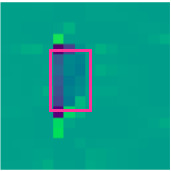}
        \includegraphics[width=\curvefigwidth]{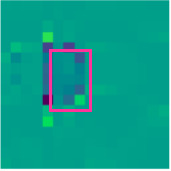}
      }%
    \end{minipage}
    \caption{\small{\textbf{Illustration of GFM Effect}. We visualize the gradient maps of five consecutive video frames by EP~\cite{extremalPerturbations} and our proposed F-EP method. Red boxes are foreground objects. They clearly show that the gradients of our method are more spatiotemporal consistent.}}
    \label{fig:centroid}
    \vspace{-4mm}
\end{figure*}

\newcommand{\figwidth}{0.1\textwidth}
\begin{figure*}
\setlength{\abovecaptionskip}{0.0cm}
\setlength{\belowcaptionskip}{0.1cm}
\setlength{\tabcolsep}{3mm}
    \begin{minipage}[b]{1.\linewidth}
      \subfigure[Original, CricketBowling, 0.59]{
        \includegraphics[width=\figwidth]{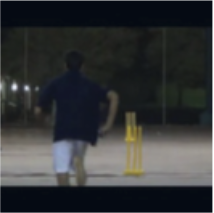}
        \includegraphics[width=\figwidth]{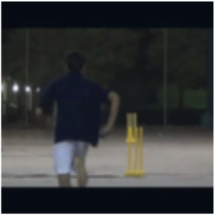}
        \includegraphics[width=\figwidth]{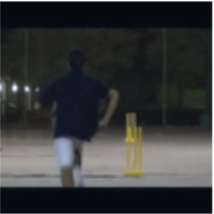}
        }
      \centering
      \subfigure[EP \cite{extremalPerturbations}, CricketBowling, 1.0]{
        \includegraphics[width=\figwidth]{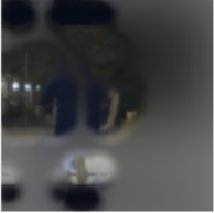}
        \includegraphics[width=\figwidth]{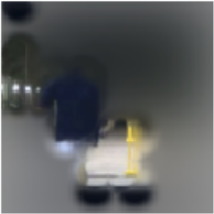}
        \includegraphics[width=\figwidth]{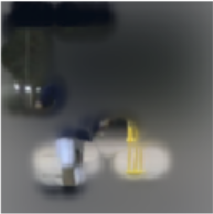}
        }
      \subfigure[F-EP, CricketBowling, 0.85]{
        \includegraphics[width=\figwidth]{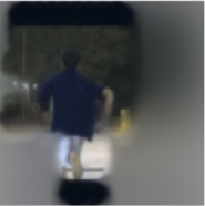}
        \includegraphics[width=\figwidth]{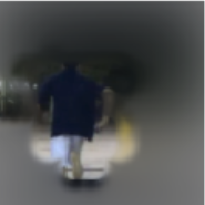}
        \includegraphics[width=\figwidth]{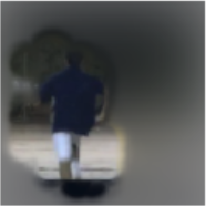}
        }
    \end{minipage}
    \caption{\small{\textbf{Interpretability vs Faithfulness}. Under each figure is the method, predicted class and probability. Without GFM, EP \cite{extremalPerturbations} produces explanations that are noisy and not spatiotemporal consistent with attention on the background and part of human motion. While F-EP gives more interpretable explanations, the faithfulness can sometimes be compromised, i.e., accuracy of 0.85 vs 1.0.}}
    \label{fig:noise_vs_int}
    \vspace{-4mm}
\end{figure*}

\paragraphX{Motivation.} A model explanation is expected to be interpretable by people and faithfully represent a model's inference process~\cite{xai}. This becomes a challenge when the neural network not only uses low-frequency features for learning larger shapes and objects like faces, but also high-frequency features that look like noise~\cite{high-frequency}. Because EP aims to achieve higher model confidence regardless of the the frequencies chosen, thus the high frequency features are included if they are more salient than the low frequency features which make explanations noisy and not spatiotemporally consistent. 

Because the gradients encode the importance of each pixel (Eq.~\eqref{eq:gradients-ep}), important high-frequency features would have high gradients in $\nabla M$, making $\nabla M$ noisy. A noisy $\nabla M$ leads to noisy explanations $M$, since the latter is iteratively updated with the former, as shown in Eq.~\eqref{eq:gradient-ascent}. We propose to transform the gradients into the frequency domain using the Discrete Cosine Transform (DCT), such that the signals related to the content-defining features, which reside in the lower end of the spectrum of the frequencies~\cite{jpeg}, are preserved. High-frequency signals are not only associated with noise, but also the fine-grained details such as textures/edges. Thus, the low frequency signals allow the explanations to focus on the target object/person and become spatiotemporal consistent, while the high frequency features make the explanations more faithful to the model and fine-grained, see Fig.~\ref{fig:noise_vs_int}. The Fig. \ref{fig:centroid} shows that the frequency modulated gradient maps focus more on the foreground object and become spatiotemporal consistent when consecutive frames contain the target object. In the following section, we propose our method Frequency-based Extremal Perturbation (F-EP) and theoretically demonstrate that modulating gradients is equivalent to modulating the masks.

\paragraphX{Gradient Frequency Modulation.} F-EP is presented in Fig.~\ref{fig:f-ep}. At each iteration of the optimization, F-EP transforms the gradient maps $\nabla M$ into the frequency domain using DCT. DCT is chosen over Discrete Fourier Transform (DFT) because the basis functions used in DFT are complex-valued, while $\nabla M$ only contains real numbers, meaning that DFT would compute useless transformations~\cite{dctVSdft}. Because $\nabla M$ is three dimensional, the DCT is also three dimensional. Denote the DCT as $\mathcal{H}$ and the frequency map of the gradient $\nabla M$ as $G\in \mathbb{R}^{H\times W\times T}$, the $(i,j,k)$-th spatio-temporal entry of $G$ $(i\in\{1,\ldots,H\}$, $j\in\{1,\ldots,W\}$, and $k\in\{1,\ldots,T\})$ is computed by the three-dimensional DCT:
\begin{equation}
 \begin{split}
     G_{i,j,k} = \mathcal{H}(\nabla M)_{i,j,k} & = a \sum_{z=0}^{T-1} \sum_{y=0}^{W-1}\sum_{x=0}^{H-1}   c_x c_y c_z \nabla M_{x,y,z}d_{x, i}^{(H)} d_{y, j}^{(W)} d_{z, k}^{(T)} \\
     & = a h_k^T\left(h_j^T\left(h_{i}^T \nabla M\right)\right)
   \label{eq:dct}
 \end{split}
\end{equation}
where $a = \left(\frac{2}{H} \right)^{\frac{1}{2}} \left( \frac{2}{W} \right)^{\frac{1}{2}}\left( \frac{2}{T} \right)^{\frac{1}{2}}$ is a constant and $d_{x,i}^{(H)} = cos \left [(2x+1)i\pi/(2H)\right]$ is the cosine basis function, $c_x=1/\sqrt{2}$ if $x=0$, otherwise $c_x=1$. $h_i\in \mathbb{R}^{H\times 1}$ is the vector form of the element-wise product between the vector $c_x$ and $d_{x,i}^{(H)}$ along $x$-axis, i.e., $h_{i} = c_x\odot d_{x,i}^{(H)}$. The definitions of $\{c_y, c_z\}$, $\{d_{y,j}^{(W)}, d_{z,k}^{(T)}\}$, and $\{h_j, h_k\}$ are similar to $c_x$, $d_{x,i}^{(H)}$, and $h_i$. Thus, Eq.~\eqref{eq:dct} shows that the DCT operation is an intrinsically linear system, and according to the gradient ascent rule in Eq.~\eqref{eq:gradient-ascent}, we have the following equality:
\begin{equation}
    \mathcal{H}(M^{f+1}) = \mathcal{H}(M^f) + \epsilon \mathcal{H}(\nabla M).
\end{equation}
This equation shows that applying frequency modulation on the gradient map $\nabla M$ is equivalent to frequency modulation on the optimal mask $M^*$. We provide a more detailed derivation in the supplementary material. Therefore, it is straightforward to linearly modulate the gradient frequency map $G$. To this end, we perform the frequency modulation on $G$ as follows:
\begin{equation}
    \nabla\Tilde{M} = \Tilde{\mathcal{H}}(A \odot G) +  \Tilde{\mathcal{H}}(B \odot G),
\label{eq:freq-modul}
\end{equation}
where $ \Tilde{\mathcal{H}}$ is the inverse DCT (IDCT) function and $\odot$ is the element-wise product. Note both $\Tilde{\mathcal{H}}$ are linear systems. The low frequency mask $A\in \mathbb{R}^{H\times W\times T}$ is defined as $A_{i,j,k}=1$ if $i \leq r_l*H, j \leq r_l*W, k \leq r_l*T$, otherwise zero, and the high frequency mask $B\in \mathbb{R}^{H\times W\times T}$ is defined as $B_{i,j,k}=0$ if $i \leq (1-r_h)*H, j \leq (1-r_h)*W, k \leq (1-r_h)*T$, otherwise one. See the $A$ and $B$ matrices in Fig.~\ref{fig:f-ep}. $r_l$ and $r_h$ denote the ratios of low- and high-frequency components to be preserved, respectively. Note that $r_l, r_h \geq 0$ and $r_l + r_h \leq 1$. 

Eq.~\eqref{eq:freq-modul} first selects a ratio $r_l$ of low-end frequencies and a ratio $r_h$ of high-end frequencies from $G$, then performs IDCT on the selected high and low frequencies separately. By summing up the gradients maps modulated at different frequencies, we ensure that the gradient information contains both the low-end and high-end features.

Finally, instead of using the raw gradient $\nabla M$ in equation~\eqref{eq:gradient-ascent}, we propose to use the frequency-modulated gradient $\nabla\Tilde{M}$ such that:
\begin{equation}
    \begin{split}
        M^{f+1} = M^{f} + \epsilon \nabla\Tilde{M}.
        \label{eq:ga-dct}
    \end{split}
\end{equation}
Since DCT and IDCT are computationally efficient, our proposed modulation does not incur too much computational cost. 


\section{Experiment}

\subsection{Experimental Settings}
\paragraphX{Evaluation Metrics.}
Evaluation has always been one of the most challenging parts of studying model explainability. Due to the lack of ground truth explanations, existing methods can only be evaluated in a post hoc manner. We follow prior work in using two standard metrics that also apply to static images. First, \textbf{Drop in Confidence} (\textbf{DC}) measures the decrease in model's confidence on the explanations compared to the original input~\cite{grad-cam++} (smaller DC is better). It is represented as $\sum_i^N \max(0, y - y_e)/N$ where $y_e = \Phi(X_e)$ and $X_e = M \odot X$ as in Eq.~\eqref{eq:ep}. Second, \textbf{Accuracy} (\textbf{Acc.}) measures the model's classification accuracy on the explanations $X_e$.

In addition, since no existing evaluation metric is suitable to assess the spatiotemporal consistency of an explanation, we propose a new metric called \textbf{Spatiotemporal Consistency} (\textbf{STC}). STC calculates the overlap between the explanations and the ground truth bounding boxes. Mathematically, we define $STC = \sum_{i,j,k} ^{H, W, T} \mathds{1}_{\{O_{i,j,k} = 1, M_{i,j,k} \geq \tau \}}$, where $O \in \mathbb{R}^{T \times 1 \times H \times W}$ are the ground truth bounding boxes where $O_{i,j,k} = 1$ for all pixels inside the bounding boxes, and 0 otherwise. $\tau$ denotes the threshold of masks $M$ during evaluation because $M$ are continuous values within range $[0,1]$. Because the bounding boxes focus tightly on the foreground object/person in each frame, a more spatiotemporally consistent explanation will have higher overlap with them.

\paragraphX{Experimental Details} We evaluate F-EP against CAM-based methods: Grad-CAM \cite{grad-cam} and Grad-CAM++ \cite{grad-cam++}, and backpropagation-based methods: Gradients~\cite{gradient}, Integrated Grad~\cite{integrated} and Smooth Grad~\cite{smoothGrad}. We extend the original implementation of these methods\footnote{https://github.com/jacobgil/pytorch-grad-cam} to the video domain. EP \cite{extremalPerturbations} was extended to the video domain in STEP \cite{step} \footnote{https://github.com/shinkyo0513/Video-Visual-Explanations}. The target models to be explained are R(2+1)D \cite{r3d} and TSM \cite{tsm}.\footnote{https://github.com/open-mmlab/mmaction2} For video datasets, we use UCF101-24 \cite{ucf101} and Epic-Kitchens-Object \cite{epic-kitchens}. UCF101-24 is a video understanding dataset of 24 actions with annotated bounding boxes and Epic-Kitchens-Object records cooking activities from a first-person point of view. The hyperparameters used are the same as in STEP \cite{step}, except the step size is 13 and sigma is 23. The parameters $r_l$ and $r_h$ are selected based on the best performance on the validation set.

\subsection{Quantitative Results}

\begin{figure}[t!]
    \floatbox[{\capbeside\thisfloatsetup{capbesideposition={left,top},capbesidewidth=0.45\textwidth}}]{figure}[\FBwidth]
    {\caption{\small{DAUC of all the compared methods, which plots the model's confidence drop with respect to the percentage of pixels deleted. Note that the most salient pixels are deleted first. \textbf{Lower Area Under the Curve (AUC) is better.} F-EP has the lowest AUC compared to other methods, which indicates that the features contained in the explanations are more salient.}}
    \label{fig:dauc}}
    {\includegraphics[width=0.4\textwidth]{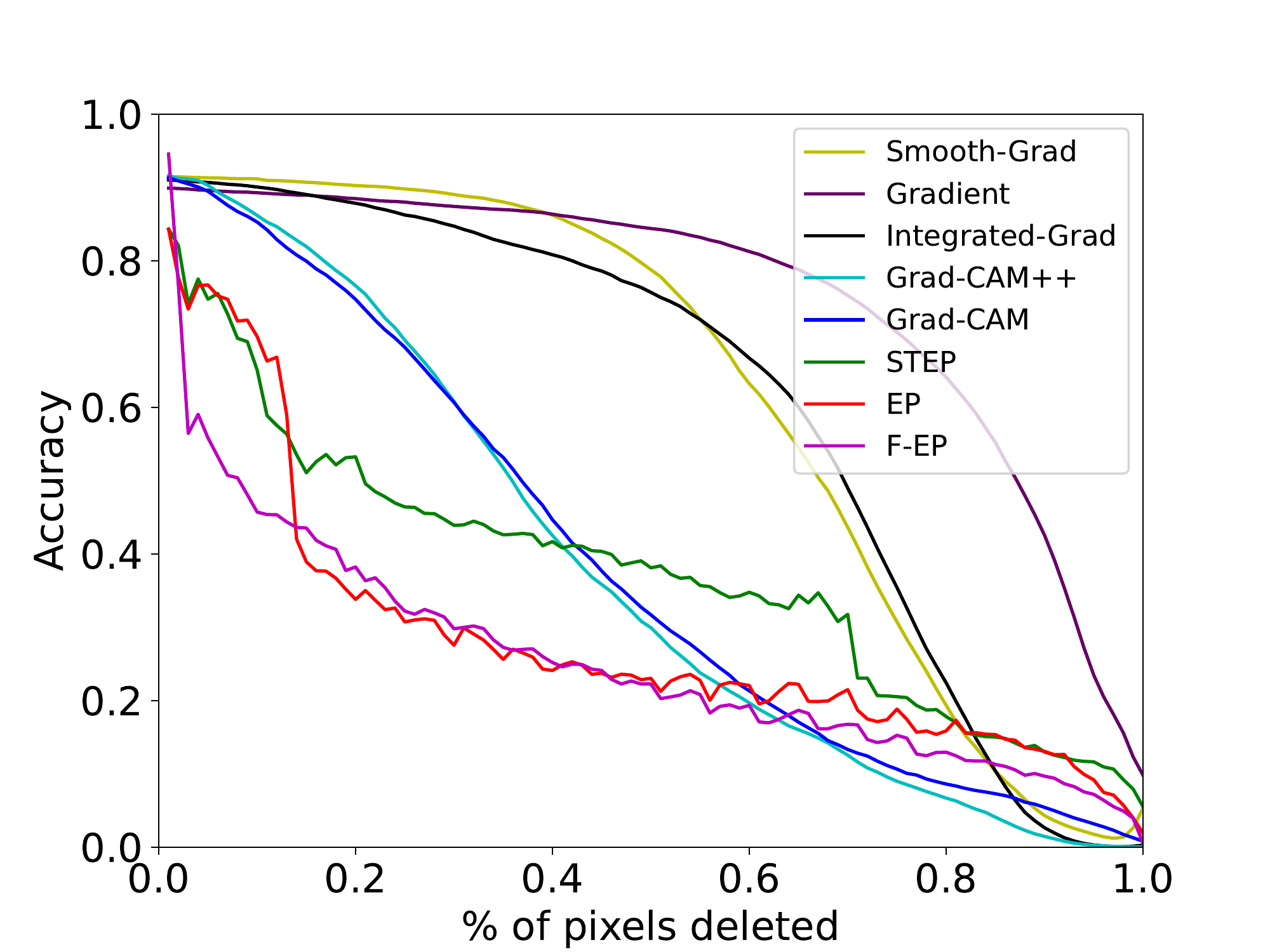}}
\end{figure}

\begin{table}
    \centering
    \footnotesize
    \captionsetup{aboveskip=2pt}
    \setlength{\tabcolsep}{2mm}
    \begin{tabular}{ l|ccc|ccc  }
    \hline
    \multirow{2}{*}{Method} &   \multicolumn{3}{c|}{R(2+1)D}  & \multicolumn{3}{c}{TSM} \\
    \cline{2-7}
                                  & DC ($\downarrow$)    & Acc. ($\uparrow$)    & STC ($\uparrow$)       & DC ($\downarrow$)     & Acc. ($\uparrow$)     & STC ($\uparrow$)     \\
     \hline
     Gradients~\cite{gradient}          &   89.4               &   6.0                &   0.4                   &   90.8                &    2.1                    &   0.6                   \\
     Integrated Grad~\cite{integrated}  &   89.5               &   6.9                &  0.5                    &  89.5                 &   4.0                     &  1.2   \\
     Smooth Grad~\cite{smoothGrad}      &   90.7               &   1.0                &  1.8                    &  90.7                 &   2.7                     &  1.2      \\
     Grad-CAM~\cite{grad-cam}           &   42.7               &   58.1               &  28.9                   &   88.6                &   11.2                    &  0.1             \\
     Grad-CAM++~\cite{grad-cam++}       &   42.3               &   58.3               &  35.1                   &\textcolor{red}{24.1}  &\textcolor{red}{74.8}      & 17.4          \\
     EP~\cite{extremalPerturbations}    &   37.9               &   61.7               &\textcolor{blue}{63.8}   &  40.0                 &   63.0                    &  70.6                       \\
     STEP~\cite{step}                   &\textcolor{blue}{34.2}   &\textcolor{blue}{67.1}&\textcolor{blue}{63.8}&  39.9                 & 65.2                      &\textcolor{blue}{73.9}       \\
     \hline
     F-EP (ours)                        &\textcolor{red}{32.9} &\textcolor{red}{73.4} &  \textcolor{red}{67.0}  &\textcolor{blue}{33.3} &\textcolor{blue}{71.0}     &  \textcolor{red}{74.6}    \\
     \hline
    \end{tabular}
    \caption{\small{Results (\%) on UCF101-24~\cite{ucf101} with R(2+1)D~\cite{r3d} and TSM~\cite{tsm} models. The best and second best performing methods are shown in \textcolor{red}{red} and \textcolor{blue}{blue}. Our method F-EP shows consistent improvement over the state-of-the-art methods on different models.}}
    \label{tab:quant_res_ucf101_r3d}
\end{table}

\paragraphX{UCF101-24 Results.}
Table \ref{tab:quant_res_ucf101_r3d} presents F-EP against SOTA methods on the UCF101-24~\cite{ucf101} dataset based on two target models, R(2+1)D~\cite{r3d} and TSM~\cite{tsm}. The outperformance of F-EP ($r_l = 0.5$, $r_h=0.2$) on the R(2+1)D model shows that it is able to produce more faithful and spatiotemporal consistent explanations than all the other methods, including the baselines EP \cite{extremalPerturbations} and STEP \cite{step}. On TSM, F-EP ($r_l=0.5$, $r_h=0.1$) achieves best on the STC metric, and second best on the DC and Acc. metrics. The first reason is that the UCF101-24 dataset has high scene representation bias~\cite{danceMall}, and the model needs not to look at the actual activity to give a correct prediction. Because the explanations given by F-EP focus more on the person performing the activity, which are at the lower end of the frequency domain, the model will have lower confidence on these explanations. Second, R(2+1)D uses 3D convolution filters to extract spatiotemporal features, while TSM uses 2D convolution filters that capture less temporal information. Thus, a more spatiotemporally consistent explanation will have higher model confidence in R(2+1)D than TSM. 

Fig. \ref{fig:dauc} illustrates the Deletion metric \cite{rise}, which measures the drop in model's confidence when the most important pixels are removed. The importance of pixels are given by the saliency maps. A lower Area Under the Curve (AUC) implies the explanation method is better and the saliency features contained are more faithful to the model. We see that F-EP has the lowest AUC compared to other explanation methods.

\begin{SCtable}
    \footnotesize
    \captionsetup{aboveskip=2pt}
    \setlength{\tabcolsep}{2mm}
    \begin{tabular}{ l|c|c|c  }
    \hline
     Method                             & DC ($\downarrow$)     & Acc. ($\uparrow$)     & STC ($\uparrow$)             \\
     \hline
     Gradients~\cite{gradient}          & 48.1                  & 16.0                      & 0.4                       \\
     Integrated Grad~\cite{integrated}  & 49.0                  & 5.5                       & 0.4                        \\
     Smooth Grad~\cite{smoothGrad}      & 49.3                  & 6.1                       & 0.8                        \\
     Grad-CAM~\cite{grad-cam}           &\textcolor{red}{28.3}  &\textcolor{red}{48.0}      & 27.2                         \\
     Grad-CAM++~\cite{grad-cam++}       & 54.8                  & 42.1                      & 30.6                  \\
     EP~\cite{extremalPerturbations}    & 34.2                  & 43.4                      & 58.0                  \\
     STEP~\cite{step}                   & 32.6                  & 43.8                      &\textcolor{blue}{61.0}           \\
     \hline
     F-EP (ours)                        &\textcolor{blue}{32.4} &\textcolor{blue}{44.4}      &\textcolor{red}{67.8}           \\
     \hline
    \end{tabular}
    \caption{\small{Results (\%) on Epic-Kitchens~\cite{epic-kitchens} with R(2+1)D~\cite{r3d}. The best and second performance are shown in \textcolor{red}{red} and \textcolor{blue}{blue}. The explanations given by F-EP are more spatiotemporal consistent than SOTA method while being comparable in the metrics of faithfulness (DC and Acc.).}}
    \label{tab:quant_res_epic_r3d}
\end{SCtable}

\paragraphX{Epic-Kitchens-Objects Results.} Table~\ref{tab:quant_res_epic_r3d} reports the results of R(2+1)D on the Epic-Kitchens-Object~\cite{epic-kitchens} dataset. F-EP ($r_l=0.7$, $r_h=0$) performs slightly less well on DC and Acc than Grad-CAM, but much better in STC (Qualitative results are shown in the supplemental material). The explanations given by F-EP have lower model confidence because videos in the Epic-Kitchens-Object dataset do not contain the target object in every frame but F-EP produces explanation for each frame. Therefore, non-target objects are included in the explanations which have lower model confidence. A future research direction is to have the optimization algorithm attend only to salient frames for the output decision. 

Tables~\ref{tab:quant_res_ucf101_r3d} and \ref{tab:quant_res_epic_r3d} show that backpropagation-based methods, Gradients~\cite{gradient}, Integrated Grad~\cite{integrated} and SmoothGrad~\cite{smoothGrad}, perform poorly compared to CAM- and perturbation-based methods. For example, Fig.~\ref{fig:comparison_vis}'s second row (more results in the supplement) shows that Integrated Grad's explanations consist of sparse pixels that align poorly with the ground truth. These explanations are often perceived by the model as adversarial samples and have lower model confidence or even incorrect predicted labels~\cite{ian}. SmoothGrad and Gradients produce similar explanations and hence face the same problems.

\begin{figure}
    \floatbox[{\capbeside\thisfloatsetup{capbesideposition={left,top},capbesidewidth=0.35\textwidth}}]{figure}[\FBwidth]
    {\caption{\small{Performance comparison of different $r_l$ and $r_h$. 
    The vertical series of circles show the performance of $r_l = \{0.3, 0.4, 0.5\}$ with increasing $r_h$. Note that $r_l + r_h \leq 1$. The horizontal series of circles show when $r_h=0$ with increasing $r_l$. EP \cite{extremalPerturbations} is the lower right circle.
    }}
    \label{fig:ablation_studies}}
    {\includegraphics[width=0.4\textwidth]{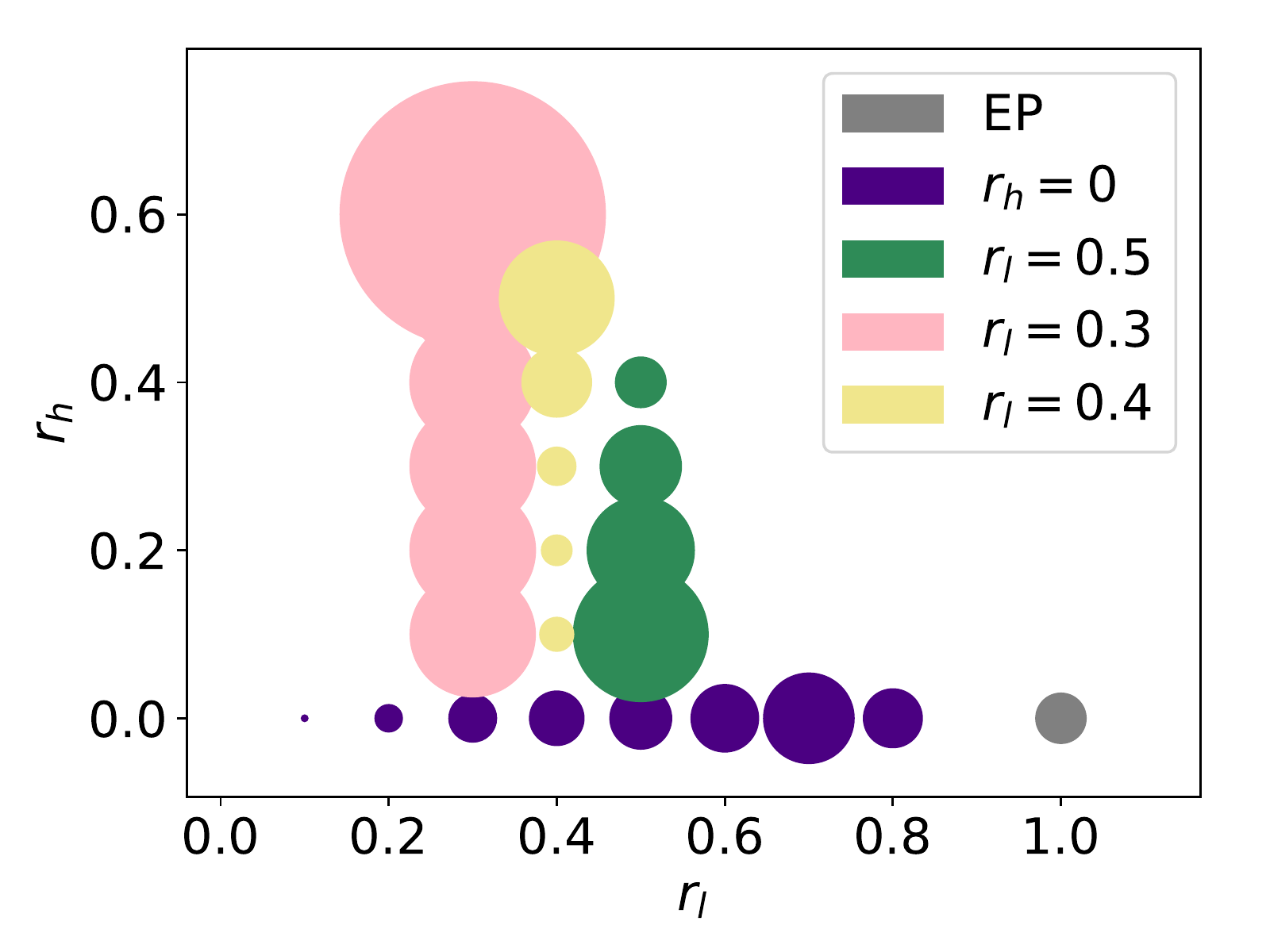}}
\end{figure}

\squeezeup

\subsection{Ablation Studies} 

\newcommand{\framewidth}{0.13\linewidth}
\newcolumntype{T}{>{\tiny}l}
\newcolumntype{H}{>{\Huge}l}

\begin{figure*}[t]
\footnotesize
\centering
\renewcommand{\tabcolsep}{0.7pt} %
\begin{tabular}{>{\scriptsize}ccccccccc}
\parbox[c]{4mm}{\multirow{1}{*}[3.0em]{\rotatebox[origin=c]{90}{Input}}} &
\includegraphics[width=\framewidth]{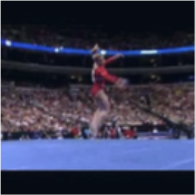} &
\includegraphics[width=\framewidth]{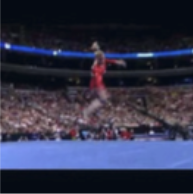} &
\includegraphics[width=\framewidth]{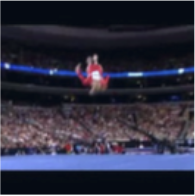} &
\includegraphics[width=\framewidth]{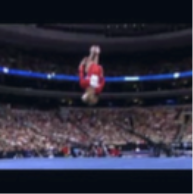} &
\includegraphics[width=\framewidth]{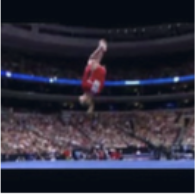} &
\includegraphics[width=\framewidth]{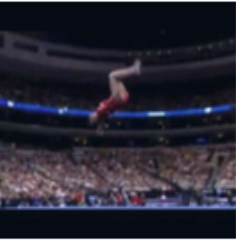} &
\includegraphics[width=\framewidth]{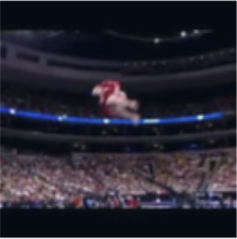} &

\parbox[c]{4mm}{\multirow{1}{*}[3.5em]{\rotatebox[origin=c]{90}{FG, [0.63]}}}
\\
\parbox[c]{4mm}{\multirow{1}{*}[5.5em]{\rotatebox[origin=c]{90}{Grad-CAM~\cite{grad-cam}}}} &
\includegraphics[width=\framewidth]{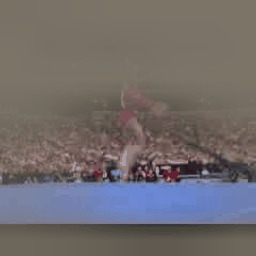} &
\includegraphics[width=\framewidth]{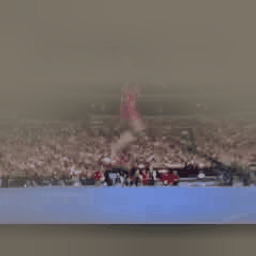} &
\includegraphics[width=\framewidth]{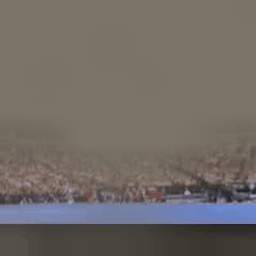} &
\includegraphics[width=\framewidth]{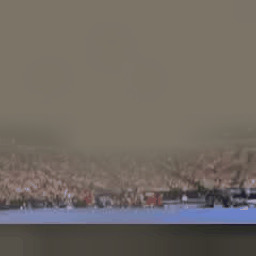} &
\includegraphics[width=\framewidth]{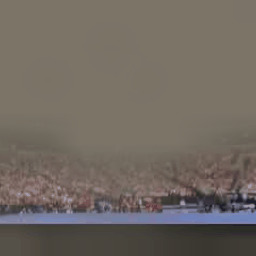} &
\includegraphics[width=\framewidth]{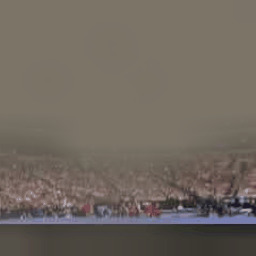} &
\includegraphics[width=\framewidth]{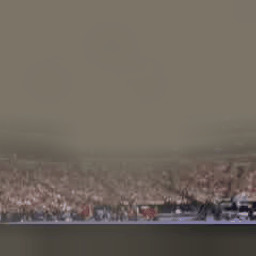} &

\parbox[c]{4mm}{\multirow{1}{*}[3.5em]{\rotatebox[origin=c]{90}{BB, [0.19]}}} 
\\
\parbox[c]{4mm}{\multirow{1}{*}[4.5em]{\rotatebox[origin=c]{90}{Integrated~\cite{integrated}}}} &
\includegraphics[width=\framewidth]{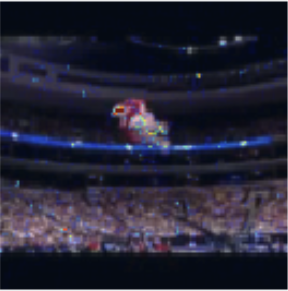} &
\includegraphics[width=\framewidth]{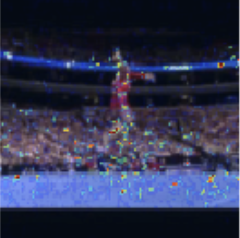} &
\includegraphics[width=\framewidth]{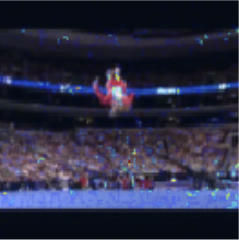} &
\includegraphics[width=\framewidth]{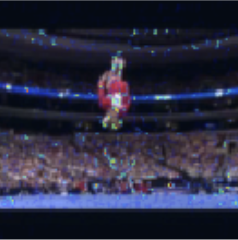} &
\includegraphics[width=\framewidth]{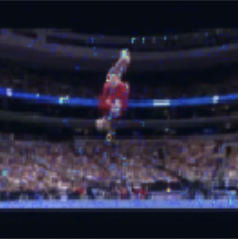} &
\includegraphics[width=\framewidth]{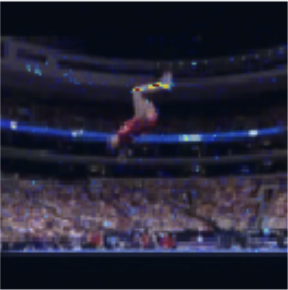} &
\includegraphics[width=\framewidth]{images/r3d_ucf101/r3d_ucf101_integrated_grad/floorGymnastics_g01_c01_49-64-12.png} &
\parbox[c]{4mm}{\multirow{1}{*}[3.5em]{\rotatebox[origin=c]{90}{WB, [0.37]}}} 
\\
\parbox[t]{4mm}{\multirow{1}{*}[4.0em]{\rotatebox[origin=c]{90}{EP~\cite{extremalPerturbations}}}} &
\includegraphics[width=\framewidth]{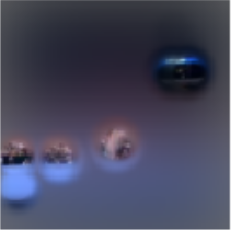} &
\includegraphics[width=\framewidth]{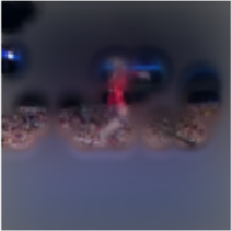} &
\includegraphics[width=\framewidth]{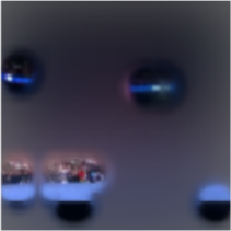} &
\includegraphics[width=\framewidth]{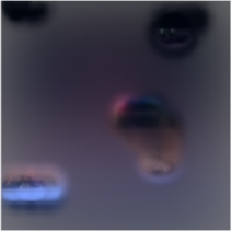} &
\includegraphics[width=\framewidth]{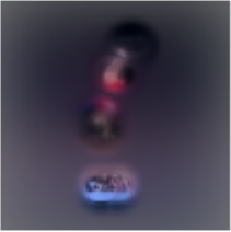} &
\includegraphics[width=\framewidth]{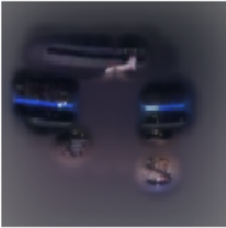} &
\includegraphics[width=\framewidth]{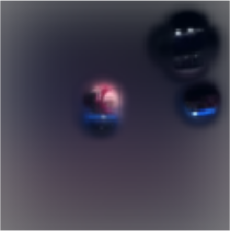} &
\parbox[c]{4mm}{\multirow{1}{*}[3.5em]{\rotatebox[origin=c]{90}{BB, [0.99]}}} 
\\
\parbox[t]{4mm}{\multirow{1}{*}[4.0em]{\rotatebox[origin=c]{90}{STEP~\cite{step}}}} &
\includegraphics[width=\framewidth]{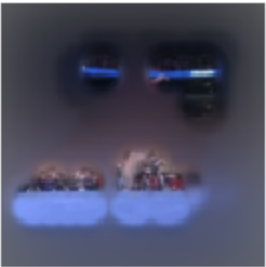} &
\includegraphics[width=\framewidth]{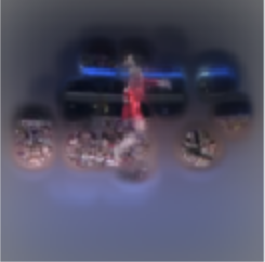} &
\includegraphics[width=\framewidth]{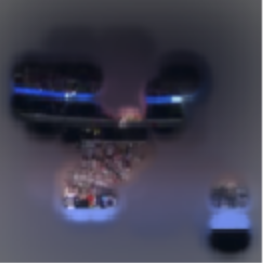} &
\includegraphics[width=\framewidth]{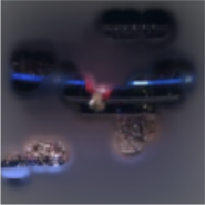} &
\includegraphics[width=\framewidth]{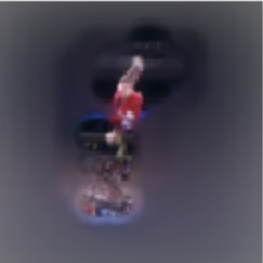} &
\includegraphics[width=\framewidth]{images/r3d_ucf101/r3d_ucf101_step/floorGymnastics_g01_c01_49-64-10.png} &
\includegraphics[width=\framewidth]{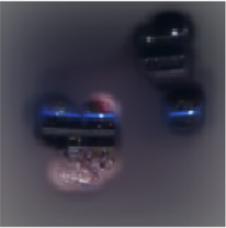} &
\parbox[c]{4mm}{\multirow{1}{*}[3.5em]{\rotatebox[origin=c]{90}{BB, [0.87]}}} 
\\
\parbox[t]{4mm}{\multirow{1}{*}[4.0em]{\rotatebox[origin=c]{90}{F-EP (ours)}}} &
\includegraphics[width=\framewidth]{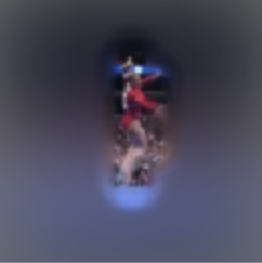} &
\includegraphics[width=\framewidth]{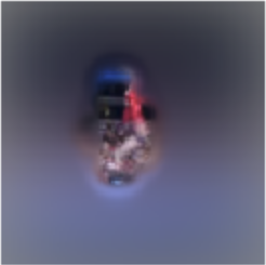} &
\includegraphics[width=\framewidth]{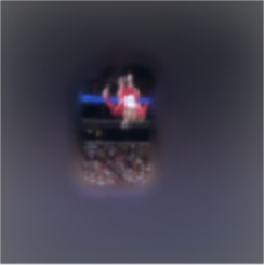} &
\includegraphics[width=\framewidth]{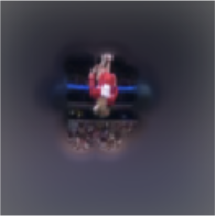} &
\includegraphics[width=\framewidth]{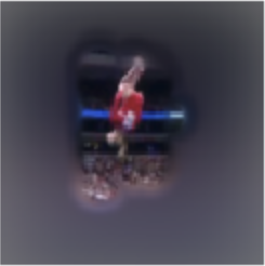} &
\includegraphics[width=\framewidth]{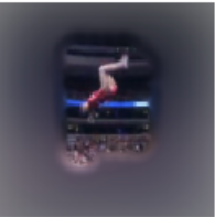} &
\includegraphics[width=\framewidth]{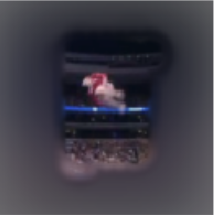} &
\parbox[c]{4mm}{\multirow{1}{*}[3.5em]{\rotatebox[origin=c]{90}{FG, [0.76]}}} 
\\
\end{tabular}
\captionsetup{font=small,aboveskip=3pt}
\caption{\textbf{Visual comparison of explanation methods on the UCF101-24 dataset ~\cite{ucf101}.} The input (first row) contains consecutive frames from the activity FloorGymnastics. On the right of each method, the first word is the predicted label on the explanation where FG = FloorGymnastics, BB = BalanceBeam, WB = WritingOnBoard, and the number denotes the predicted probability.
}
\label{fig:comparison_vis}
\vspace{-10pt}
\end{figure*}

\vspace{-5pt}
\paragraphX{Without High Frequency.} In Fig.~\ref{fig:ablation_studies}, when $r_h=0$ and as $r_l$ increases, the STC performance increases because the higher frequency signals encode the details such as edges/textures to increase faithfulness of explanations. There is a sharp drop in performance after $r_l=0.7$, because as $\nabla M$ contains more and more higher frequency signals, it is overwhelmed with the excess amount of details. The explanations also become noisier and less spatiotemporal consistent, because higher frequency features are preferred over the low frequency features if they could increase more model confidence.

\vspace{-5pt}
\paragraphX{Combination of High and Low Frequencies.} We conduct experiments with varying $r_h$ and deterministic $r_l = \{0.4, 0.5, 0.6\}$. When $r_l=0.3$, as $r_h$ increases, STC increases because the amount of frequency signals on the lower end frequency spectrum is not enough and higher frequency signals add valuable information to the explanations. However, when $r_l=0.4$, STC performance increases less quickly than $r_l=0.3$. This could mean that $r_l = 0.4$ is a threshold at which the utility of high frequency signals decreases. Finally, when $r_l = 0.5$, STC actually decreases when $r_h$ increases. Also, F-EP is able to outperform the baseline method EP at multiple combination of frequencies which confirms our hypothesis that not all frequency signals are salient. One future work direction is to search the appropriate ratios $r_l$ and $r_h$ for each sample because samples vary across dataset and actions.

\subsection{Qualitative Results}

In Fig.~\ref{fig:comparison_vis}, we see that the explanations produced by F-EP are much easier to interpret than those of the prior works. Integrated Grad explanations are noisy, sparse and difficult to interpret, because it directly uses gradients, which contain a substantial amount of noise. Grad-CAM produces explanations that fail to capture the activity dynamics. EP and STEP has difficulty in correctly capturing the person's movement in the consecutive frames. In comparison, F-EP is able to produce explanations that focus on the person performing the activity on each frame (more qualitative results are in the supplement).

\section{Conclusion}

In this paper, we propose an explanation method for video understanding models called Frequency-based Extremal Perturbation (F-EP) which aims to produce spatiotemporal consistent explanations that are faithful to the model and interpretable by humans. F-EP transforms the gradients to the frequency domain and modulates the frequency signals in order to preserve those gradient components that help to explain a model's decision. We experiment F-EP on various datasets and models and show that F-EP is able to outperform existing state-of-the-art works. 

\paragraphX{Acknowledgments.} This work is supported in part by the John S. and James L. Knight Foundation, the Miami Foundation through the Ethics and Governance of the Artificial Intelligence Initiative, the National Science Foundation under award No. 2040209, and the Army Research Office under grant number W911NF-21-1-0236.

\bibliography{egbib}

\begin{thebibliography}{48}
\providecommand{\natexlab}[1]{#1}
\providecommand{\url}[1]{\texttt{#1}}
\expandafter\ifx\csname urlstyle\endcsname\relax
  \providecommand{\doi}[1]{doi: #1}\else
  \providecommand{\doi}{doi: \begingroup \urlstyle{rm}\Url}\fi

\bibitem[Adebayo et~al.(2018)Adebayo, Gilmer, Muelly, Goodfellow, Hardt, and
  Kim]{sanity-checks}
Julius Adebayo, Justin Gilmer, Michael Muelly, Ian Goodfellow, Moritz Hardt,
  and Been Kim.
\newblock Sanity checks for saliency maps.
\newblock In \emph{NeurIPS}, 2018.

\bibitem[Adel~Bargal et~al.(2018)Adel~Bargal, Zunino, Kim, Zhang, Murino, and
  Sclaroff]{cebr}
Sarah Adel~Bargal, Andrea Zunino, Donghyun Kim, Jianming Zhang, Vittorio
  Murino, and Stan Sclaroff.
\newblock Excitation backprop for rnns.
\newblock In \emph{CVPR}, 2018.

\bibitem[Blinn(1993)]{dctVSdft}
J.F. Blinn.
\newblock What's that deal with the dct?
\newblock \emph{IEEE Computer Graphics and Applications}, 13, 1993.

\bibitem[Chattopadhay et~al.(2018)Chattopadhay, Sarkar, Howlader, and
  Balasubramanian]{grad-cam++}
Aditya Chattopadhay, Anirban Sarkar, Prantik Howlader, and Vineeth~N
  Balasubramanian.
\newblock Grad-cam++: Generalized gradient-based visual explanations for deep
  convolutional networks.
\newblock In \emph{WACV}, 2018.

\bibitem[Cho et~al.(2014)Cho, van Merri{\"e}nboer, Bahdanau, and
  Bengio]{machine_trans}
Kyunghyun Cho, Bart van Merri{\"e}nboer, Dzmitry Bahdanau, and Yoshua Bengio.
\newblock On the properties of neural machine translation: Encoder{--}decoder
  approaches.
\newblock In \emph{Proceedings of {SSST}-8, Eighth Workshop on Syntax,
  Semantics and Structure in Statistical Translation}, pages 103--111, Doha,
  Qatar, October 2014. Association for Computational Linguistics.
\newblock \doi{10.3115/v1/W14-4012}.
\newblock URL \url{https://www.aclweb.org/anthology/W14-4012}.

\bibitem[Choi et~al.(2019)Choi, Gao, Messou, and Huang]{danceMall}
Jinwoo Choi, Chen Gao, C.~E.~Joseph Messou, and Jia-Bin Huang.
\newblock Why can't i dance in the mall? learning to mitigate scene bias in
  action recognition.
\newblock In \emph{NeurIPS}, 2019.

\bibitem[Damen et~al.(2018)Damen, Doughty, Farinella, Fidler, Furnari, Kazakos,
  Moltisanti, Munro, Perrett, Price, and Wray]{epic-kitchens}
Dima Damen, Hazel Doughty, Giovanni~Maria Farinella, Sanja Fidler, Antonino
  Furnari, Evangelos Kazakos, Davide Moltisanti, Jonathan Munro, Toby Perrett,
  Will Price, and Michael Wray.
\newblock Scaling egocentric vision: The epic-kitchens dataset.
\newblock In \emph{ECCV}, 2018.

\bibitem[{Deng} et~al.(2009){Deng}, {Dong}, {Socher}, {Li}, {Kai Li}, and {Li
  Fei-Fei}]{imagenet}
J.~{Deng}, W.~{Dong}, R.~{Socher}, L.~{Li}, {Kai Li}, and {Li Fei-Fei}.
\newblock Imagenet: A large-scale hierarchical image database.
\newblock In \emph{CVPR}, 2009.

\bibitem[Desai and Ramaswamy(2020)]{ablation-cam}
Saurabh Desai and Harish~G. Ramaswamy.
\newblock Ablation-cam: Visual explanations for deep convolutional network via
  gradient-free localization.
\newblock In \emph{WACV}, 2020.

\bibitem[Dzindolet et~al.(2003)Dzindolet, Peterson, Pomranky, Pierce, and
  Beck]{ml-1}
Mary~T. Dzindolet, Scott~A. Peterson, Regina~A. Pomranky, Linda~G. Pierce, and
  Hall~P. Beck.
\newblock The role of trust in automation reliance.
\newblock \emph{Int. J. Hum.-Comput. Stud.}, 58, 2003.

\bibitem[Fong et~al.(2019)Fong, Patrick, and Vedaldi]{extremalPerturbations}
Ruth Fong, Mandela Patrick, and Andrea Vedaldi.
\newblock Understanding deep networks via extremal perturbations and smooth
  masks.
\newblock In \emph{ICCV}, 2019.

\bibitem[Fong and Vedaldi(2017)]{meaningful}
Ruth~C. Fong and Andrea Vedaldi.
\newblock Interpretable explanations of black boxes by meaningful perturbation.
\newblock In \emph{ICCV}, 2017.

\bibitem[Fu et~al.(2020)Fu, Hu, Dong, Guo, Gao, and Li]{xgrad-cam}
Ruigang Fu, Qingyong Hu, Xiaohu Dong, Yulan Guo, Yinghui Gao, and Biao Li.
\newblock Axiom-based grad-cam: Towards accurate visualization and explanation
  of cnns.
\newblock In \emph{BMVC}, 2020.

\bibitem[Gu et~al.(2019)Gu, Yang, and Tresp]{clrp}
Jindong Gu, Yinchong Yang, and Volker Tresp.
\newblock Understanding individual decisions of cnns via contrastive
  backpropagation.
\newblock In \emph{ACCV}, 2019.

\bibitem[Gunning and Aha(2019)]{xai}
David Gunning and David Aha.
\newblock Darpa’s explainable artificial intelligence (xai) program.
\newblock \emph{AI Magazine}, 40, 2019.
\newblock URL
  \url{https://ojs.aaai.org/index.php/aimagazine/article/view/2850}.

\bibitem[Guo et~al.(2020)Guo, Frank, and Weinberger]{guo-low}
Chuan Guo, Jared~S. Frank, and Kilian~Q. Weinberger.
\newblock Low frequency adversarial perturbation.
\newblock In \emph{Proceedings of The Uncertainty in Artificial Intelligence
  Conference}, 2020.

\bibitem[Ian~Goodfellow(2014)]{ian}
Christian~Szegedy Ian~Goodfellow, Jon~Shlens.
\newblock Explaining and harnessing adversarial examples.
\newblock In \emph{ICLR}, 2014.

\bibitem[Karami et~al.(2012)Karami, Beheshti, and Yazdi]{svm-compress}
Azam Karami, Soosan Beheshti, and Mehran Yazdi.
\newblock Hyperspectral image compression using 3d discrete cosine transform
  and support vector machine learning.
\newblock In \emph{ISSPA}, 2012.

\bibitem[Li et~al.(2021)Li, Wang, Li, Huang, and Sato]{step}
Zhenqiang Li, Weimin Wang, Zuoyue Li, Yifei Huang, and Yoichi Sato.
\newblock Towards visually explaining video understanding networks with
  perturbation.
\newblock In \emph{WACV}, 2021.

\bibitem[Lin et~al.(2019)Lin, Gan, and Han]{tsm}
Ji~Lin, Chuang Gan, and Song Han.
\newblock Tsm: Temporal shift module for efficient video understanding.
\newblock In \emph{ICCV}, 2019.

\bibitem[Lipton(2018)]{ml-2}
Zachary~C. Lipton.
\newblock The mythos of model interpretability: In machine learning, the
  concept of interpretability is both important and slippery.
\newblock \emph{Queue}, 16, 2018.

\bibitem[Lombrozo(2006)]{cognitive-psychology}
Tania Lombrozo.
\newblock The structure and function of explanations.
\newblock \emph{Trends in cognitive sciences}, 10, 2006.

\bibitem[Montavon et~al.(2017)Montavon, Lapuschkin, Binder, Samek, and
  Müller]{lrp}
Grégoire Montavon, Sebastian Lapuschkin, Alexander Binder, Wojciech Samek, and
  Klaus-Robert Müller.
\newblock Explaining nonlinear classification decisions with deep taylor
  decomposition.
\newblock \emph{Pattern Recognition}, 65, 2017.

\bibitem[Petsiuk et~al.(2018)Petsiuk, Das, and Saenko]{rise}
Vitali Petsiuk, Abir Das, and Kate Saenko.
\newblock Rise: Randomized input sampling for explanation of black-box models.
\newblock In \emph{British Machine Vision Conference (BMVC)}, 2018.

\bibitem[Ribeiro et~al.(2016)Ribeiro, Singh, and Guestrin]{LIME}
Marco~Tulio Ribeiro, Sameer Singh, and Carlos Guestrin.
\newblock "why should i trust you?": Explaining the predictions of any
  classifier.
\newblock In \emph{Proceedings of the 22nd ACM SIGKDD International Conference
  on Knowledge Discovery and Data Mining}, 2016.

\bibitem[Richens et~al.(2020)Richens, Lee, and Johri]{richens2020improving}
Jonathan~G. Richens, Ciar{\~A}{\textexclamdown}n~M. Lee, and Saurabh Johri.
\newblock Improving the accuracy of medical diagnosis with causal machine
  learning.
\newblock \emph{Nature Communications}, 11, 2020.

\bibitem[Roy et~al.(2017)Roy, Chakraborty, Sameer, and Naskar]{dct-class}
Aniket Roy, Rajat~Subhra Chakraborty, Udaya Sameer, and Ruchira Naskar.
\newblock Camera source identification using discrete cosine transform residue
  features and ensemble classifier.
\newblock In \emph{CVPRW}, 2017.

\bibitem[Selvaraju et~al.(2017)Selvaraju, Cogswell, Das, Vedantam, Parikh, and
  Batra]{grad-cam}
Ramprasaath~R Selvaraju, Michael Cogswell, Abhishek Das, Ramakrishna Vedantam,
  Devi Parikh, and Dhruv Batra.
\newblock Grad-cam: Visual explanations from deep networks via gradient-based
  localization.
\newblock In \emph{ICCV}, 2017.

\bibitem[Sharma et~al.(2019)Sharma, Ding, and Brubaker]{low-frequency}
Yash Sharma, Gavin~Weiguang Ding, and Marcus~A. Brubaker.
\newblock On the effectiveness of low frequency perturbations.
\newblock In \emph{IJCAI}, 2019.

\bibitem[Shen et~al.(2019)Shen, Margolies, Rothstein, Fluder, McBride, and
  Sieh]{tumor_detection}
Li~Shen, Laurie~R. Margolies, Joseph~H. Rothstein, Eugene Fluder, Russell
  McBride, and Weiva Sieh.
\newblock Deep learning to improve breast cancer detection on screening
  mammography.
\newblock \emph{Scientific Reports}, 9, 2019.

\bibitem[Simonyan et~al.(2014)Simonyan, Vedaldi, and Zisserman]{gradient}
Karen Simonyan, Andrea Vedaldi, and Andrew Zisserman.
\newblock Deep inside convolutional networks: Visualising image classification
  models and saliency maps.
\newblock In \emph{ICLR Workshop}, 2014.

\bibitem[Sixt et~al.(2019)Sixt, Granz, and Landgraf]{exp-lie}
Leon Sixt, Maximilian Granz, and Tim Landgraf.
\newblock When explanations lie: Why modified {BP} attribution fails.
\newblock In \emph{ICML}, 2019.

\bibitem[{Smilkov} et~al.(2017){Smilkov}, {Thorat}, {Kim}, {Vi{\'e}gas}, and
  {Wattenberg}]{smoothGrad}
Daniel {Smilkov}, Nikhil {Thorat}, Been {Kim}, Fernanda {Vi{\'e}gas}, and
  Martin {Wattenberg}.
\newblock {SmoothGrad: removing noise by adding noise}.
\newblock In \emph{ICML Workshop}, 2017.

\bibitem[Soomro et~al.(2012)Soomro, Zamir, Shah, Soomro, Zamir, and
  Shah]{ucf101}
Khurram Soomro, Amir~Roshan Zamir, Mubarak Shah, Khurram Soomro, Amir~Roshan
  Zamir, and Mubarak Shah.
\newblock Ucf101: A dataset of 101 human actions classes from videos in the
  wild.
\newblock \emph{CoRR}, abs/1212.0402, 2012.

\bibitem[Springenberg et~al.(2015)Springenberg, Dosovitskiy, Brox, and
  Riedmiller]{guided-backprop}
J.T. Springenberg, A.~Dosovitskiy, T.~Brox, and M.~Riedmiller.
\newblock Striving for simplicity: The all convolutional net.
\newblock In \emph{ICLR Workshop}, 2015.

\bibitem[Stergiou et~al.(2019)Stergiou, Kapidis, Kalliatakis, Chrysoulas,
  Veltkamp, and Poppe]{saliency-tubes}
Alexandros Stergiou, Georgios Kapidis, Grigorios Kalliatakis, Christos
  Chrysoulas, Remco Veltkamp, and Ronald Poppe.
\newblock Saliency tubes: Visual explanations for spatio-temporal convolutions.
\newblock In \emph{ICIP}, 2019.

\bibitem[Sundararajan et~al.(2017)Sundararajan, Taly, and Yan]{integrated}
Mukund Sundararajan, Ankur Taly, and Qiqi Yan.
\newblock Axiomatic attribution for deep networks.
\newblock In \emph{ICML}, 2017.

\bibitem[Tania(2011)]{philosophy}
Lombrozo Tania.
\newblock The instrumental value of explanations.
\newblock \emph{Philosophy Compass}, 6, 2011.

\bibitem[Tran et~al.(2014)Tran, Bourdev, Fergus, Torresani, and Paluri]{c3d}
Du~Tran, Lubomir Bourdev, Rob Fergus, Lorenzo Torresani, and Manohar Paluri.
\newblock Learning spatiotemporal features with 3d convolutional networks.
\newblock In \emph{ICCV}, 2014.

\bibitem[Tran et~al.(2018)Tran, Wang, Torresani, Ray, LeCun, and Paluri]{r3d}
Du~Tran, Heng Wang, Lorenzo Torresani, Jamie Ray, Yann LeCun, and Manohar
  Paluri.
\newblock A closer look at spatiotemporal convolutions for action recognition.
\newblock In \emph{CVPR}, 2018.

\bibitem[Verhelst et~al.(2020)Verhelst, Stannat, and Mecacci]{detection}
H.~M. Verhelst, A.~W. Stannat, and G.~Mecacci.
\newblock Machine learning against terrorism: How big data collection and
  analysis influences the privacy-security dilemma.
\newblock \emph{Science and Engineering Ethics}, 26, 2020.

\bibitem[Wagner et~al.(2019)Wagner, Kohler, Gindele, Hetzel, Wiedemer, and
  Behnke]{FGVis}
Jorg Wagner, Jan~Mathias Kohler, Tobias Gindele, Leon Hetzel, Jakob~Thaddaus
  Wiedemer, and Sven Behnke.
\newblock Interpretable and fine-grained visual explanations for convolutional
  neural networks.
\newblock In \emph{CVPR}, 2019.

\bibitem[Wallace(1992)]{jpeg}
G.K. Wallace.
\newblock The jpeg still picture compression standard.
\newblock \emph{IEEE Transactions on Consumer Electronics}, 38, 1992.

\bibitem[Wang et~al.(2020{\natexlab{a}})Wang, Du, Yang, and Zhang]{score-cam}
Haofan Wang, Mengnan Du, Fan Yang, and Zijian Zhang.
\newblock Score-cam: Improved visual explanations via score-weighted class
  activation mapping.
\newblock In \emph{CVPRW}, 2020{\natexlab{a}}.

\bibitem[Wang et~al.(2020{\natexlab{b}})Wang, Wu, Yin, and
  Xing]{high-frequency}
Haohan Wang, Xindi Wu, Pengcheng Yin, and Eric~P. Xing.
\newblock High frequency component helps explain the generalization of
  convolutional neural networks.
\newblock In \emph{CVPR}, 2020{\natexlab{b}}.

\bibitem[Zeiler and Fergus(2014)]{deconvnet}
Matthew~D. Zeiler and Rob Fergus.
\newblock Visualizing and understanding convolutional networks.
\newblock In \emph{ECCV}, 2014.

\bibitem[Zhang et~al.(2017)Zhang, Lin, Brandt, Shen, and Sclaroff]{excitation}
Jianming Zhang, Zhe Lin, Jonathan Brandt, Xiaohui Shen, and Stan Sclaroff.
\newblock Top-down neural attention by excitation backprop.
\newblock \emph{IJCV}, 126, 2017.

\bibitem[Zhou et~al.(2016)Zhou, Khosla, Lapedriza, Oliva, and Torralba]{cam}
Bolei Zhou, Aditya Khosla, Agata Lapedriza, Aude Oliva, and Antonio Torralba.
\newblock Learning deep features for discriminative localization.
\newblock In \emph{CVPR}, 2016.

\end{thebibliography}
\end{document}


\maketitle

\section{Optimality of GFM}
Below, we demonstrate that applying frequency modulation on $\nabla M$ is equivalent to the frequency modulation on the optimal mask $M^*$. We modulate the gradient map in the frequency domain, $G = \mathcal{H}(\nabla M)$, as follows:
\begin{equation}
    \begin{split}
        \nabla\Tilde{M} = \Tilde{\mathcal{H}}(\mathcal{O}_A(G)) +  \Tilde{\mathcal{H}}(\mathcal{O}_B(G)),
    \end{split}
\end{equation}
Where $\mathcal{O}_A(G) = A \odot G$ and $\mathcal{O}_B(G) = B \odot G$ are linear. We now show that the Inverse Discrete Cosine Transform, $\Tilde{\mathcal{H}}$, is also linear. $\Tilde{\mathcal{H}}$ is defined as follows:
\begin{equation}
    \begin{split}
        \Tilde{\mathcal{H}}(G)_{x,y,z} & = a \sum_{k=0}^{T-1} \sum_{j=0}^{W-1}\sum_{i=0}^{H-1}  c_i c_j c_k G_{i,j,k}d_{i, x}^{(H)} d_{j, y}^{(W)} d_{k, z}^{(T)} \\
        & = a h_z^T\left(h_y^T\left(h_{x}^T G\right)\right) \\
        \label{eq:idct}
    \end{split}
\end{equation}
The equation~\eqref{eq:idct} is a matrix multiplication and thus linear. Note that the functions $d(\cdot, \cdot)$ and $c$ are defined the same as in the paper. We also know that DCT is invertible and linear, thus IDCT is also linear \cite{dctVSdft}.

Therefore, the gradient frequency modulation can be summarized as:
\begin{equation}
    \begin{split}
        \nabla \Tilde{M} = \Tilde{\mathcal{H}} \circ \mathcal{O} \circ \mathcal{H} (\nabla M)
    \end{split}
\end{equation}
Because $\Tilde{\mathcal{H}}$, $\mathcal{H}$ and $\mathcal{O}$ are linear transformations, we have the following gradient ascent rule:
\begin{equation}
    \begin{split}
        \mathcal{F}(M^{f+1}) &= \mathcal{F}(M^f + \epsilon \nabla M) \\
        & = \mathcal{F}(M^f) + \epsilon \mathcal{F}(\nabla M)
        \label{eq:f-ga}
    \end{split}
\end{equation}
The equation~\eqref{eq:f-ga} shows that applying frequency modulation on the gradient map $\nabla M$ is equivalent to frequency modulation on the optimal mask $M^*$.

\section{Additional Ablation Studies Results}

\begin{table}[!htp]
\centering
\footnotesize
\begin{tabular}{|c|c c c c c c c c|}
\hline
$r_l$ & 0.1 & 0.2 & 0.3 & 0.4 & 0.5 & 0.6 & 0.7 & 0.8\\
\hline
STC & 55.5 & 58.8 & 61.9 & 61.9 & 62.9  & 62.7 & \textbf{67.8} & 63.5 \\
\hline
\end{tabular}
\caption{\textbf{Without high frequencies results with model R(2+1)d}. The results are the quantitative results are the purple circles in the figure 3 of the paper.}
\label{tab:low-freq-only}
\end{table}

\begin{table}[!htp]\centering
\footnotesize
\begin{tabular}{|l|c c c c c c | c c c c c|}
\hline
$r_l$ &0.3 &0.3 &0.3 &0.3 &0.3 &0.3 &0.4 &0.4 &0.4 &0.4 &0.4 \\
\hline
$r_h$ &0.1 &0.2 &0.3 &0.4 &0.5 &0.6 &0.1 &0.2 &0.3 &0.4 &0.5 \\
\hline
STC & 61.9 & 61.9 & 61.9 & 61.9 & 61.9 & 62.7 & 61.9 & 61.8 & 62.0 & 62.8 & 62.8 \\
\hline
\end{tabular}
\caption{\textbf{Low and High Frequencies where $r_l = \{0.3, 0.4\}$}. This table presents the results of the using a combination of low and high frequencies on the dataset Epic-Kitchens-Object and the model R(2+1)D.}
\label{tab:comb-1}

\end{table}

\begin{table}[!htp]\centering
\footnotesize
\begin{tabular}{|l| c c c c | c c c |}
\toprule
\hline
$r_l$ &0.5 &0.5 &0.5 &0.5 &0.6 &0.6 &0.6 \\
\hline
$r_h$ &0.1 &0.2 &0.3 &0.4 &0.1 &0.2 &0.3 \\
\hline
STC & 62.9 & 62.5 & 62.1 & 61.7 & 62.7 & \textbf{63.0} & 62.7 \\
\hline
\bottomrule
\end{tabular}
\caption{\textbf{Low and High Frequencies where $r_l = \{0.5, 0.6\}$}. This table presents the results of the using a combination of low and high frequencies on the dataset Epic-Kitchens-Object and the model R(2+1)D.}
\label{tab:comp-2}
\end{table}

In table \ref{tab:low-freq-only}, we present the results of the STC performance using only low frequencies. Note that the table corresponds to the horizontal purple circles in the figure 3 of the paper. The table \ref{tab:comb-1} reports the performance using a combination of low and high frequency signals which correspond to the vertical pink and yellow circles in the figure 3. The table \ref{tab:comp-2} reports the results of $r_l = \{0.5, 0.6\}$ where $r_l = \{0.5\}$ are the green circles in the figure 3. 

We see that when the performance of F-EP using a combination of low and high frequencies is superior than the baselines (3D\_EP has 58.0, STEP has 61.0), but lower than the best performance of using low frequency signals only which is 67.8 at $r_l = 0.7$. For $r_l = \{0.3, 0.4\}$, a larger amount of high frequency signals improve the STC performance, while when $r_l$ increases, $r_l = \{0.5, 0.6\}$, an increasing amount of high frequency signals do not necessarily help for better spatiotemporal consistency.

\section{Qualitative Results}
Figure \ref{fig:ucf-tsm} compares the baseline methods with F-EP on the dataset UCF101-24 and the model TSM. F-EP is able to localize the activity which is biking accurately both spatially and temporally. Although Grad-CAM and Grad-CAM++ also localize the biking girl, a substantial amount of background information is also included. Backpropagation-based methods Gradients, Integrated Grad and SmoothGrad produce noisy explanations that are hard to interpret.

Figure \ref{fig:r3d_epic} compares the methods on the dataset Epic-Kitchens-Object and the model R(2+1)D. We see that F-EP is able to localize the object cupboard for most of the scene, but it also highlights the closet which is not the target object (4th frame). One future work direction is to optimize the masks to attend only to salient frames. CAM- and backpropagation-based methods are noisy and not spatiotemporal consistent, e.g., Grad-CAM++ and Grad-CAM fail to localize the cupboard. The baselines 3D\_EP and STEP fails to localize cupboard in the last two frames and the cupboard at the second object is masked by the explanations. Integrated Grad and SmoothGrad do not capture the cupboard object at the second frame.

\newcommand{\framewidth}{0.13\linewidth}
\newcolumntype{T}{>{\tiny}l}
\newcolumntype{H}{>{\Huge}l}

\begin{figure*}[t]
\footnotesize
\centering
\renewcommand{\tabcolsep}{1.5pt} %
\begin{tabular}{>{\scriptsize}cccccccc}
\parbox[c]{4mm}{\multirow{1}{*}[3.0em]{\rotatebox[origin=c]{90}{Input}}} &
\includegraphics[width=\framewidth]{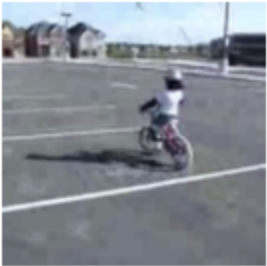} &
\includegraphics[width=\framewidth]{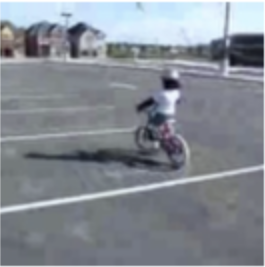} &
\includegraphics[width=\framewidth]{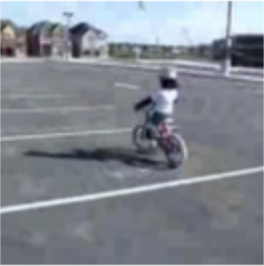} &
\includegraphics[width=\framewidth]{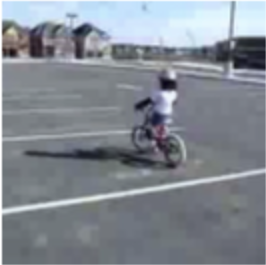} &
\includegraphics[width=\framewidth]{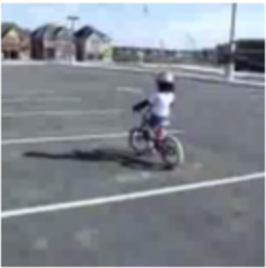} &
\includegraphics[width=\framewidth]{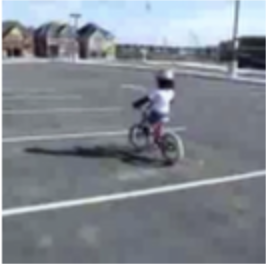} &
\parbox[c]{4mm}{\multirow{1}{*}[3.5em]{\rotatebox[origin=c]{90}{BK, [0.926]}}}
\\
\parbox[c]{4mm}{\multirow{1}{*}[5.5em]{\rotatebox[origin=c]{90}{Grad-CAM++\cite{grad-cam++}}}} &
\includegraphics[width=\framewidth]{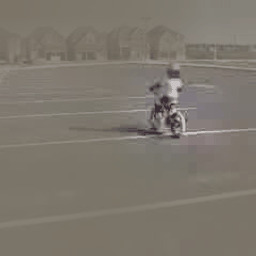} &
\includegraphics[width=\framewidth]{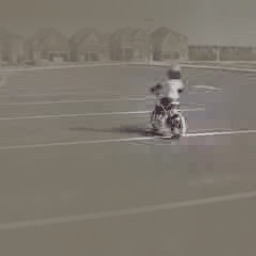} &
\includegraphics[width=\framewidth]{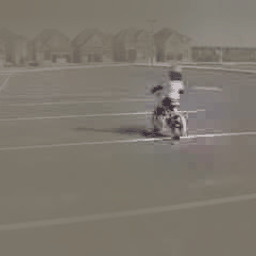} &
\includegraphics[width=\framewidth]{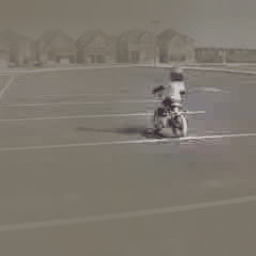} &
\includegraphics[width=\framewidth]{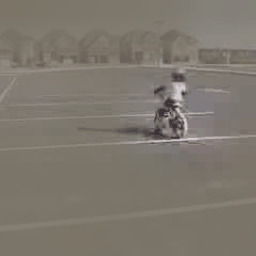} &
\includegraphics[width=\framewidth]{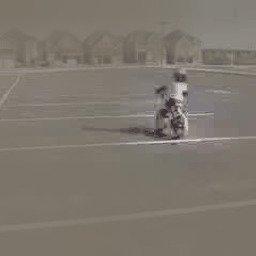} &

\parbox[c]{4mm}{\multirow{1}{*}[3.5em]{\rotatebox[origin=c]{90}{BK, [0.4201]}}}
\\
\parbox[c]{4mm}{\multirow{1}{*}[5.5em]{\rotatebox[origin=c]{90}{Grad-CAM\cite{grad-cam}}}} &
\includegraphics[width=\framewidth]{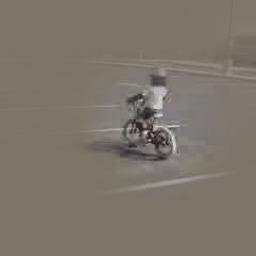} &
\includegraphics[width=\framewidth]{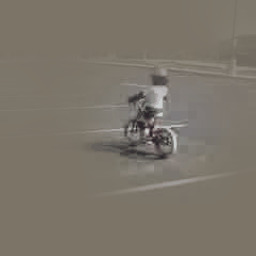} &
\includegraphics[width=\framewidth]{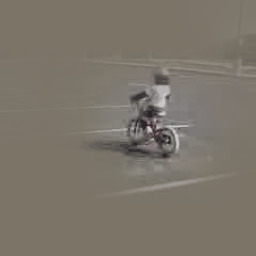} &
\includegraphics[width=\framewidth]{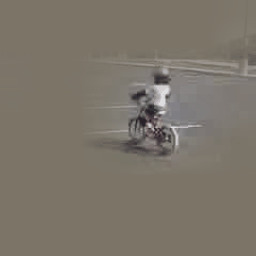} &
\includegraphics[width=\framewidth]{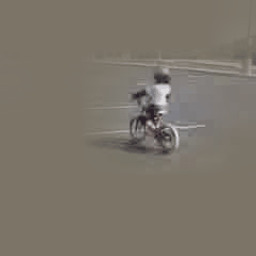} &
\includegraphics[width=\framewidth]{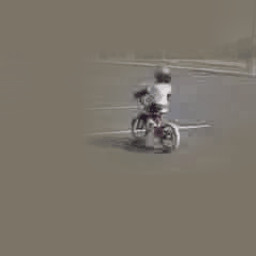} &

\parbox[c]{4mm}{\multirow{1}{*}[3.5em]{\rotatebox[origin=c]{90}{CD, [0.0023]}}} 
\\
\parbox[c]{4mm}{\multirow{1}{*}[4.5em]{\rotatebox[origin=c]{90}{Integrated\cite{integrated}}}} &
\includegraphics[width=\framewidth]{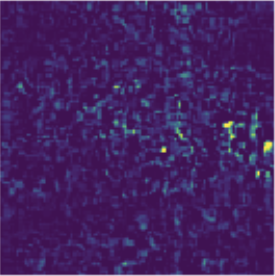} &
\includegraphics[width=\framewidth]{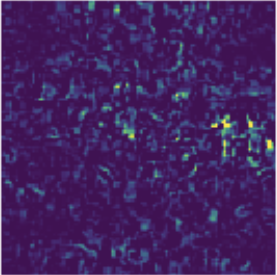} &
\includegraphics[width=\framewidth]{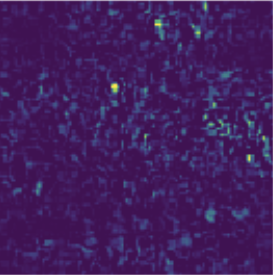} &
\includegraphics[width=\framewidth]{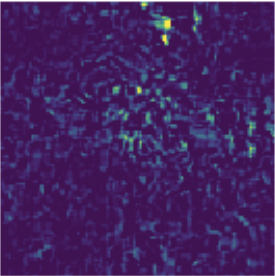} &
\includegraphics[width=\framewidth]{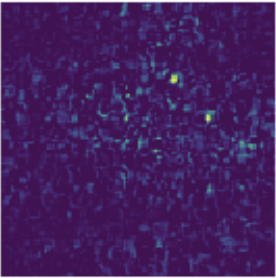} &
\includegraphics[width=\framewidth]{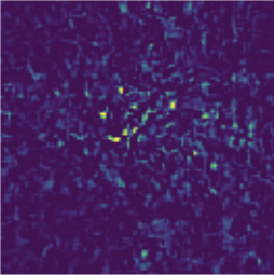} &

\parbox[c]{4mm}{\multirow{1}{*}[3.5em]{\rotatebox[origin=c]{90}{CD, [0.0013]}}} 
\\
\parbox[c]{4mm}{\multirow{1}{*}[4.5em]{\rotatebox[origin=c]{90}{Gradients\cite{gradient}}}} &
\includegraphics[width=\framewidth]{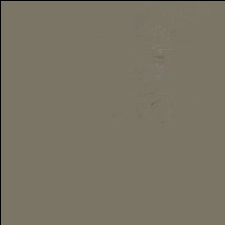} &
\includegraphics[width=\framewidth]{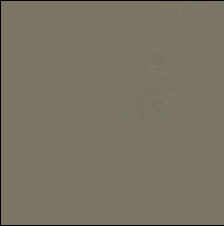} &
\includegraphics[width=\framewidth]{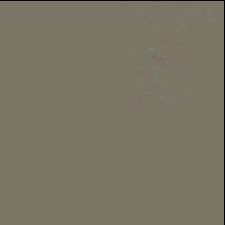} &
\includegraphics[width=\framewidth]{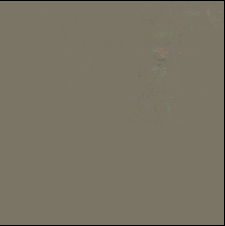} &
\includegraphics[width=\framewidth]{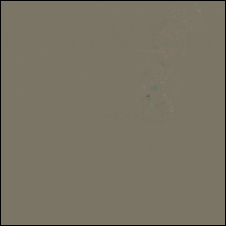} &
\includegraphics[width=\framewidth]{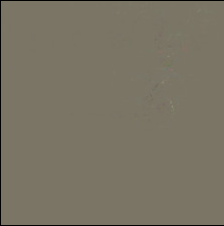} &

\parbox[c]{4mm}{\multirow{1}{*}[3.5em]{\rotatebox[origin=c]{90}{CD, [0.001]}}} 
\\
\parbox[c]{4mm}{\multirow{1}{*}[4.5em]{\rotatebox[origin=c]{90}{SmoothGrad\cite{smoothGrad}}}} &
\includegraphics[width=\framewidth]{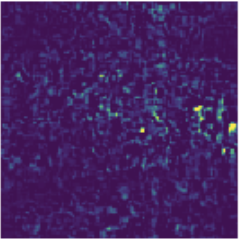} &
\includegraphics[width=\framewidth]{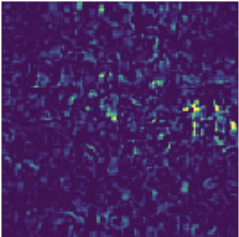} &
\includegraphics[width=\framewidth]{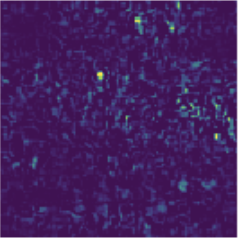} &
\includegraphics[width=\framewidth]{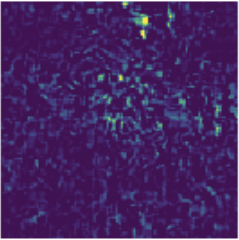} &
\includegraphics[width=\framewidth]{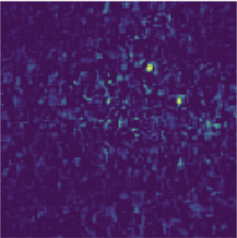} &
\includegraphics[width=\framewidth]{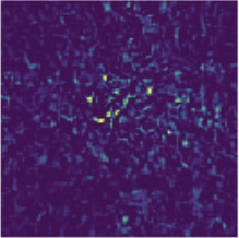} &

\parbox[c]{4mm}{\multirow{1}{*}[3.5em]{\rotatebox[origin=c]{90}{RCI, [0.084]}}} 
\\
\parbox[t]{4mm}{\multirow{1}{*}[4.0em]{\rotatebox[origin=c]{90}{EP\cite{extremalPerturbations}}}} &
\includegraphics[width=\framewidth]{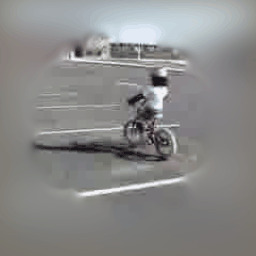} &
\includegraphics[width=\framewidth]{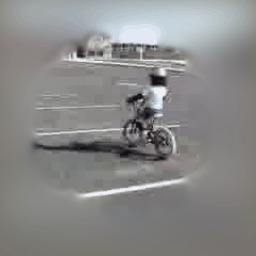} &
\includegraphics[width=\framewidth]{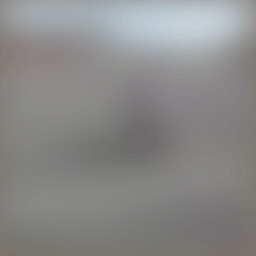} &
\includegraphics[width=\framewidth]{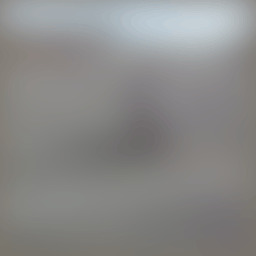} &
\includegraphics[width=\framewidth]{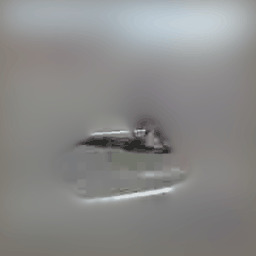} &
\includegraphics[width=\framewidth]{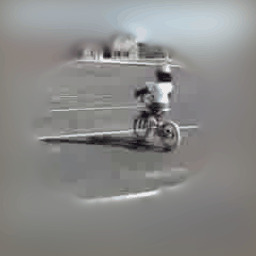} &

\parbox[c]{4mm}{\multirow{1}{*}[3.5em]{\rotatebox[origin=c]{90}{BK, [0.5]}}} 
\\
\parbox[t]{4mm}{\multirow{1}{*}[4.0em]{\rotatebox[origin=c]{90}{STEP\cite{step}}}} &
\includegraphics[width=\framewidth]{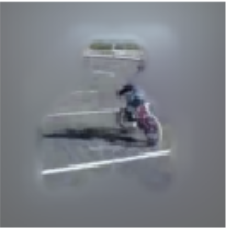} &
\includegraphics[width=\framewidth]{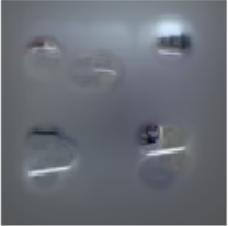} &
\includegraphics[width=\framewidth]{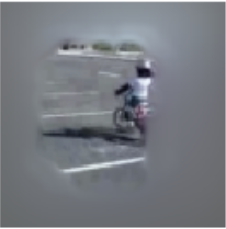} &
\includegraphics[width=\framewidth]{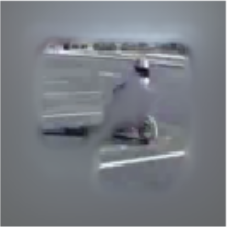} &
\includegraphics[width=\framewidth]{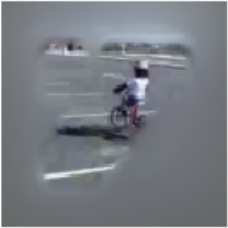} &
\includegraphics[width=\framewidth]{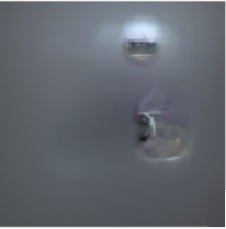} &
\parbox[c]{4mm}{\multirow{1}{*}[3.5em]{\rotatebox[origin=c]{90}{BK, [0.99]}}} 
\\
\parbox[t]{4mm}{\multirow{1}{*}[4.0em]{\rotatebox[origin=c]{90}{F-EP (ours)}}} &
\includegraphics[width=\framewidth]{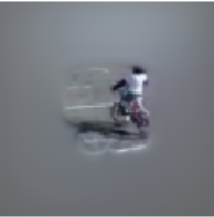} &
\includegraphics[width=\framewidth]{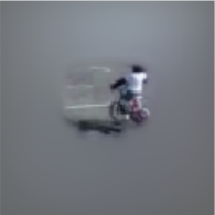} &
\includegraphics[width=\framewidth]{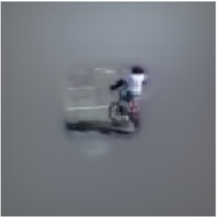} &
\includegraphics[width=\framewidth]{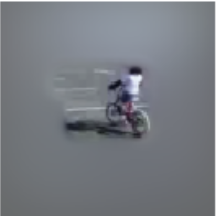} &
\includegraphics[width=\framewidth]{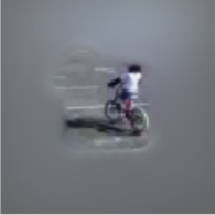} &
\includegraphics[width=\framewidth]{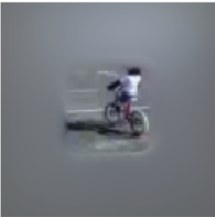} &
\parbox[c]{4mm}{\multirow{1}{*}[3.5em]{\rotatebox[origin=c]{90}{BK, [0.97]}}} 
\\
\end{tabular}
\captionsetup{font=small,aboveskip=3pt}
\caption{\textbf{Visual comparison of explanation methods on the UCF101-24 dataset ~\cite{ucf101} with model TSM \cite{tsm}}. The input (first row) contains consecutive frames from the activity Biking. On the right of each method, the first word is the predicted label on the explanation where BK = Biking, CD = CliffDiving, RCI = RockClimbingIndoor, and the number denotes the predicted probability.
}
\label{fig:ucf-tsm}
\vspace{-10pt}
\end{figure*}

\clearpage

\begin{figure*}[t]
\footnotesize
\centering
\renewcommand{\tabcolsep}{1.5pt} %
\begin{tabular}{>{\scriptsize}cccccccc}
\parbox[c]{4mm}{\multirow{1}{*}[3.0em]{\rotatebox[origin=c]{90}{Input}}} &
\includegraphics[width=\framewidth]{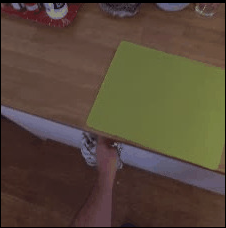} &
\includegraphics[width=\framewidth]{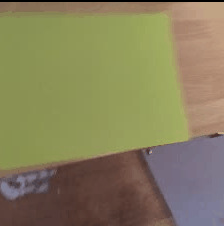} &
\includegraphics[width=\framewidth]{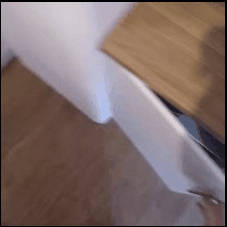} &
\includegraphics[width=\framewidth]{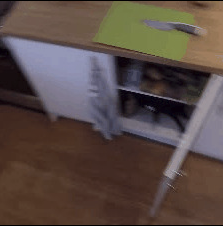} &
\includegraphics[width=\framewidth]{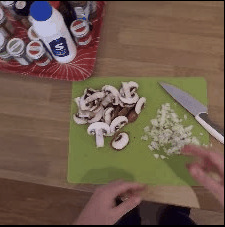} &
\includegraphics[width=\framewidth]{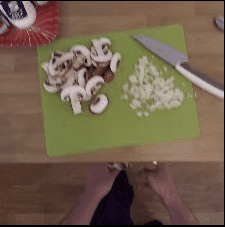} &

\parbox[c]{4mm}{\multirow{1}{*}[3.5em]{\rotatebox[origin=c]{90}{CB, [0.20]}}}
\\
\parbox[c]{4mm}{\multirow{1}{*}[5.5em]{\rotatebox[origin=c]{90}{Grad-CAM\cite{grad-cam}}}} &
\includegraphics[width=\framewidth]{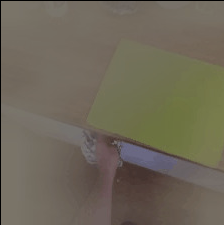} &
\includegraphics[width=\framewidth]{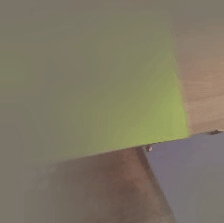} &
\includegraphics[width=\framewidth]{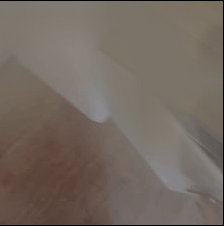} &
\includegraphics[width=\framewidth]{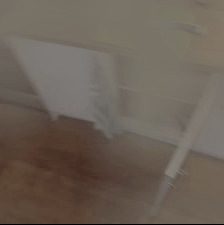} &
\includegraphics[width=\framewidth]{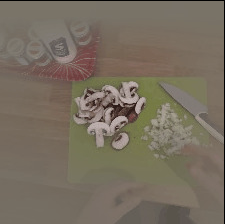} &
\includegraphics[width=\framewidth]{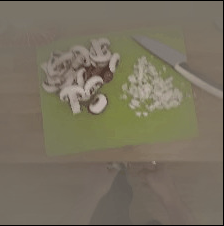} &

\parbox[c]{4mm}{\multirow{1}{*}[3.5em]{\rotatebox[origin=c]{90}{Tap, [0.10]}}}
\\
\parbox[c]{4mm}{\multirow{1}{*}[5.5em]{\rotatebox[origin=c]{90}{Grad-CAM++\cite{grad-cam++}}}} &
\includegraphics[width=\framewidth]{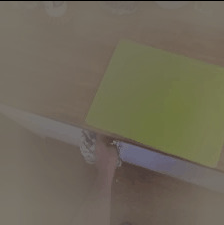} &
\includegraphics[width=\framewidth]{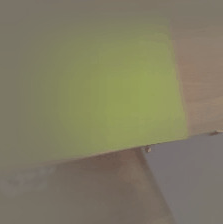} &
\includegraphics[width=\framewidth]{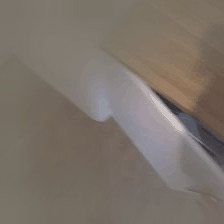} &
\includegraphics[width=\framewidth]{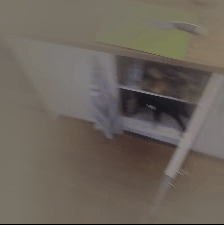} &
\includegraphics[width=\framewidth]{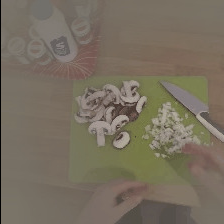} &
\includegraphics[width=\framewidth]{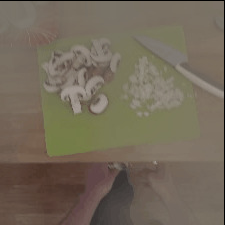} &

\parbox[c]{4mm}{\multirow{1}{*}[3.5em]{\rotatebox[origin=c]{90}{CB, [0.10]}}} 
\\
\parbox[c]{4mm}{\multirow{1}{*}[4.5em]{\rotatebox[origin=c]{90}{Integrated\cite{integrated}}}} &
\includegraphics[width=\framewidth]{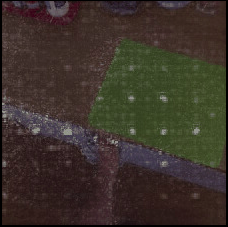} &
\includegraphics[width=\framewidth]{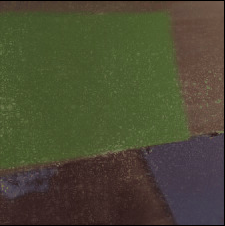} &
\includegraphics[width=\framewidth]{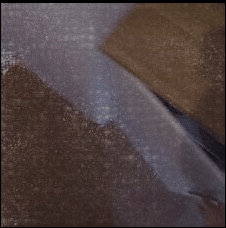} &
\includegraphics[width=\framewidth]{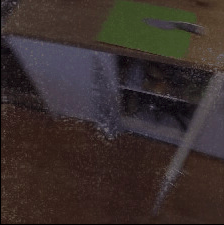} &
\includegraphics[width=\framewidth]{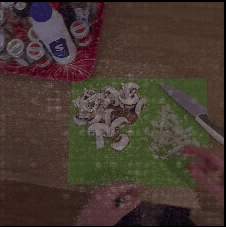} &
\includegraphics[width=\framewidth]{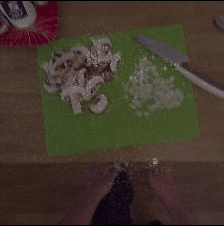} &

\parbox[c]{4mm}{\multirow{1}{*}[3.5em]{\rotatebox[origin=c]{90}{Tap, [0.03]}}} 
\\
\parbox[c]{4mm}{\multirow{1}{*}[4.5em]{\rotatebox[origin=c]{90}{Gradient\cite{gradient}}}} &
\includegraphics[width=\framewidth]{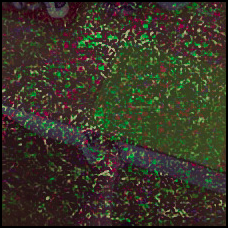} &
\includegraphics[width=\framewidth]{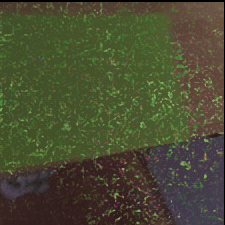} &
\includegraphics[width=\framewidth]{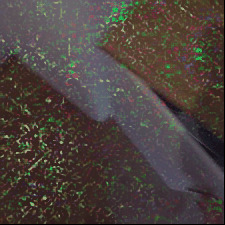} &
\includegraphics[width=\framewidth]{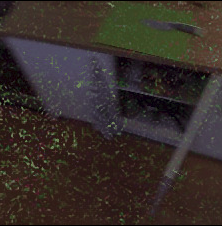} &
\includegraphics[width=\framewidth]{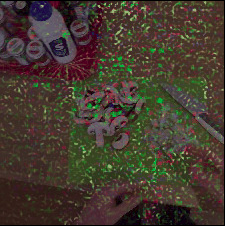} &
\includegraphics[width=\framewidth]{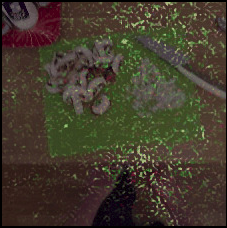} &

\parbox[c]{4mm}{\multirow{1}{*}[3.5em]{\rotatebox[origin=c]{90}{Tap, [0.03]}}} 
\\
\parbox[c]{4mm}{\multirow{1}{*}[4.5em]{\rotatebox[origin=c]{90}{SmoothGrad\cite{smoothGrad}}}} &
\includegraphics[width=\framewidth]{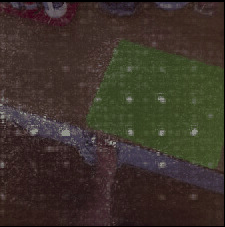} &
\includegraphics[width=\framewidth]{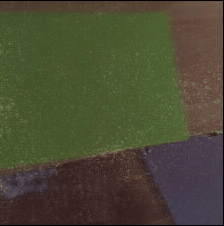} &
\includegraphics[width=\framewidth]{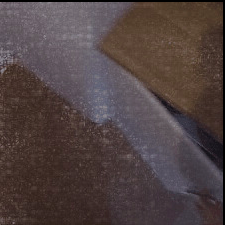} &
\includegraphics[width=\framewidth]{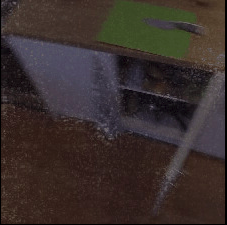} &
\includegraphics[width=\framewidth]{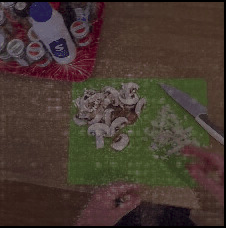} &
\includegraphics[width=\framewidth]{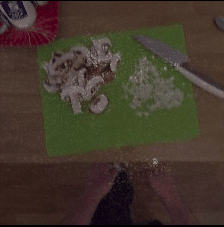} &

\parbox[c]{4mm}{\multirow{1}{*}[3.5em]{\rotatebox[origin=c]{90}{Oil, [0.04]}}} 
\\
\parbox[c]{4mm}{\multirow{1}{*}[4.5em]{\rotatebox[origin=c]{90}{3D\_EP\cite{step}}}} &
\includegraphics[width=\framewidth]{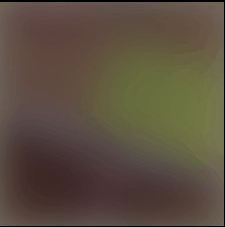} &
\includegraphics[width=\framewidth]{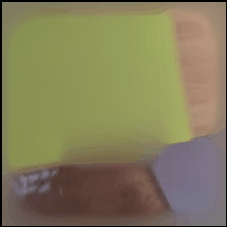} &
\includegraphics[width=\framewidth]{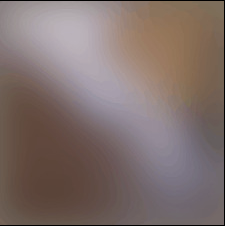} &
\includegraphics[width=\framewidth]{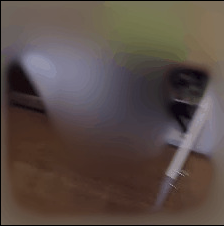} &
\includegraphics[width=\framewidth]{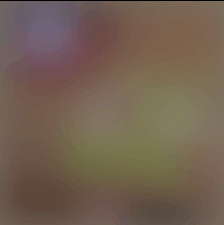} &
\includegraphics[width=\framewidth]{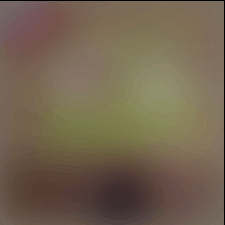} &

\parbox[c]{4mm}{\multirow{1}{*}[3.5em]{\rotatebox[origin=c]{90}{CB, [0.29]}}} 
\\
\parbox[c]{4mm}{\multirow{1}{*}[4.5em]{\rotatebox[origin=c]{90}{STEP\cite{step}}}} &
\includegraphics[width=\framewidth]{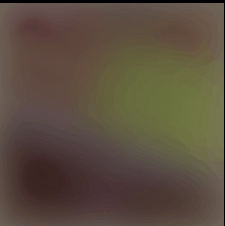} &
\includegraphics[width=\framewidth]{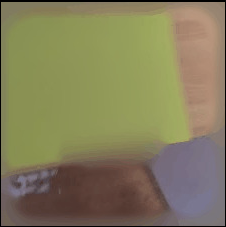} &
\includegraphics[width=\framewidth]{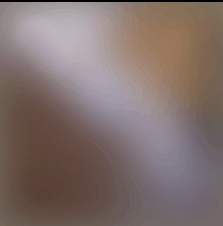} &
\includegraphics[width=\framewidth]{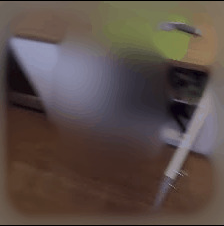} &
\includegraphics[width=\framewidth]{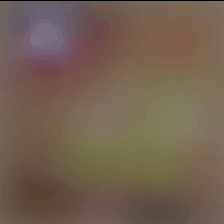} &
\includegraphics[width=\framewidth]{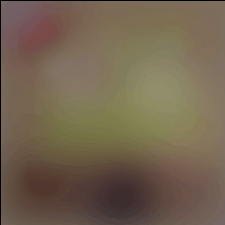} &

\parbox[c]{4mm}{\multirow{1}{*}[3.5em]{\rotatebox[origin=c]{90}{CB, [0.34]}}} 
\\
\parbox[c]{4mm}{\multirow{1}{*}[4.5em]{\rotatebox[origin=c]{90}{F-EP (Ours)}}} &
\includegraphics[width=\framewidth]{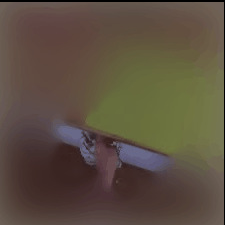} &
\includegraphics[width=\framewidth]{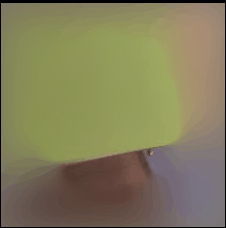} &
\includegraphics[width=\framewidth]{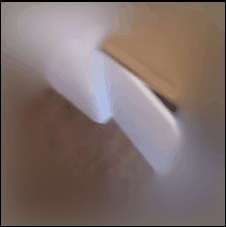} &
\includegraphics[width=\framewidth]{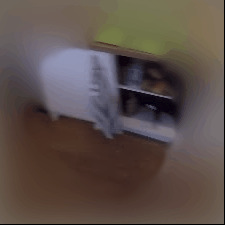} &
\includegraphics[width=\framewidth]{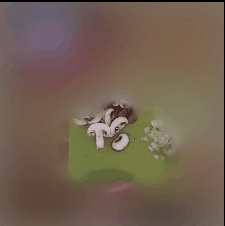} &
\includegraphics[width=\framewidth]{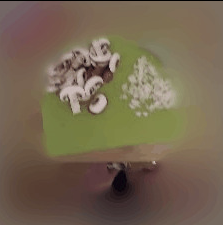} &
\parbox[c]{4mm}{\multirow{1}{*}[3.5em]{\rotatebox[origin=c]{90}{CB, [0.34]}}} 
\\
\end{tabular}
\captionsetup{font=small,aboveskip=3pt}
\caption{\textbf{Visual comparison of explanation methods on the Epic-Kitchens-Object dataset \cite{epic-kitchens}}. The input (first row) contains frames from the object cupboard. On the right of each method, the first word is the predicted label on the explanation where CB = cupboard and the number denotes the corresponding predicted probability.
}
\label{fig:r3d_epic}
\vspace{-10pt}
\end{figure*}

\clearpage
\bibliography{egbib}